# The Sigma-max System Induced from Randomness & Fuzziness and its Application in Time Series Prediction

Wei Mei*, Ming Li, Yuanzeng Cheng, Limin Liu
Army Engineering University, Shijiazhuang, 050003, P.R. China

**Abstract** This paper managed to induce probability theory (sigma system) and possibility theory (max system) respectively from the clearly-defined randomness and fuzziness, while focusing the question why the key axiom of "maxitivity" is adopted for possibility measure. Such an objective is achieved by following three steps: a) the establishment of mathematical definitions of randomness and fuzziness; b) the development of intuitive definition of possibility as measure of fuzziness based on compatibility interpretation; c) the abstraction of the axiomatic definitions of probability/ possibility from their intuitive definitions, by taking advantage of properties of the well-defined randomness and fuzziness. We derived the conclusion that "max" is the only but un-strict disjunctive operator that is applicable across the fuzzy event space, and is an exact operator for extracting the value from the fuzzy sample space that leads to the largest possibility of one. Then a demonstration example of stock price prediction is presented, which confirms that max inference indeed exhibits distinctive performance, with an improvement up to 18.99%, over sigma inference for the investigated application. Our work provides a physical foundation for the axiomatic definition of possibility for the measure of fuzziness, which hopefully would facilitate wider adoption of possibility theory in practice.

## 1. Introduction

Randomness and fuzziness are well recognized as two kinds of fundamental uncertainties of this world. It remains as an open topic on how to correctly comprehend these uncertainties and effectively handle them in practice. For modeling of random uncertainty, probability theory and the derivative subjects of statistics and stochastic process are no doubt the classic tool set. Probability theory, which satisfies the key axiom of "additivity" [16,22], has grown up to be mature, upon which nearly the whole building of information sciences is based and applications of which could be found over a great diversity of communities. For handling of fuzzy uncertainty, fuzzy sets and possibility theory stand as two alternative and related methods [40,44]. Nevertheless, it is fair to claim that their role played in information sciences is far from matching that of probability theory. Especially, the mainstream community of artificial intelligence (AI) employs, almost exclusively, probability theory to express uncertainty [14]. Fuzzy sets can mainly find its applications in fuzzy inference system, and later in fuzzy modeling and control of nonlinear systems [46]. Not like probability theory, the theory of fuzzy set is not based on the axiom system, and is not distribution-based, either. As a method parallel to probability theory, possibility theory is based on the well-known axiom of "maxitivity" [40,44] and is comfortably distribution-based. Probability and possibility are comparable because they are both based on set-functions and describe uncertainty with numbers in the unit interval [0, 1]. Possibility theory has received increasing attention in recent years by quite a few applications [23,36] but the underlying cause it outperforms probability method for these specific applications lacks a unified explanation. The reason behind this unsatisfactory situation is, recognized by some researchers, the lack of consensus on the issues pertinent to the foundation of fuzzy sets and possibility theory [14,43].

Specifically, and to the authors' knowledge, the widely acknowledged axiom of "maxitivity" has always been lack of a widely-accepted physical explanation. In contrast, the additivity axiom of probability may find its origin from the classic frequency definition of probability [16,22]. This paper is to induce probability theory and possibility theory respectively from the clearly-defined randomness and fuzziness, with a focus on answering the question why the key axiom of "maxitivity" is adopted for possibility measure. Such an objective is achieved by following three key steps: a) the establishment of mathematical definitions of randomness and fuzziness, with origin of fuzziness recognized to be closely related to the process of concept cognition; b) the development of intuitive mathematical definition of possibility as measure of fuzziness based on compatibility interpretation; c) the abstraction of the axiomatic definitions of probability/possibility from their intuitive definitions, by taking advantage of properties of the well-defined randomness and fuzziness. It should be emphasized that in the setting of this work, intuitive definitions of probability and possibility are customized respectively for the measure of randomness and fuzziness. The lack of definitions of randomness and fuzziness is leading us to a theoretical confusion when we try to develop an uncertainty theory that is distinguished from probability theory. The theoretical confusion concerns both the origins and destinations (the prospective applications) of the uncertainty theories. That is, we need to build up two different theories respectively upon two kinds of clearly-defined uncertain phenomena, to which the established uncertainty theories will eventually be applied. Compared with randomness, we would say that fuzziness may be a more complicated phenomenon, which is harder to model and carry experimental verification, because where human psychology is deeply involved.

* Corresponding author. E-mail address: meiwei@sina.com (W. Mei). ORCID: https://orcid.org/0000-0002-5736-189X (W. Mei).

Meanwhile, we also realized and as is well known, every axiomatic (abstract) theory admits of an unlimited number of concrete interpretations besides those from which it was derived [27]. For example, the de Finetti system of subjective probability established the foundations of probability theory on the notion of 'coherence', which means one should assign and manipulate probabilities so that one cannot be made a sure loser in betting based on them [13,22]. Assuming Monotonicity axiom, Anscombe-Aumann showed that if a variable is completely reversible with respect to another probabilistic variable, then it must be probabilistic too [5,14]. Giang provided a decision theoretic foundation for possibility theory using Anscombe-Aumann's approach, which says that the subjective uncertainty of a variable must obey the axioms of possibility theory if it is completely reversible with respect to the ignorant variable in Anscombe-Aumann's sense and is partially reversible with respect to the probabilistic variable [14]. All these interpretations call themselves subjective probability or subjective possibility, which could be categorized into logic probability/possibility in a broad sense since they have reference to reasonableness of belief or expectation [5]. The classic frequency definition of probability could be categorized into objective and physical probability since it has reference to a type of physical phenomenon – randomness [5]. The intuitive definition of possibility in this work should fall into the category of physical possibility since it is the measure of another type of physical phenomenon – fuzziness. We hesitate to attribute such a possibility to objective or subjective possibility because fuzziness is a kind of subjective uncertainty as well as objective physical phenomenon. Compared with other interpretations, our characterization on the foundation of possibility theory features that it is based on the intuitive definition of possibility, the measure of fuzziness.

It should be noted that possibility theory has several branches, which in our point of view are in fact different methods. The original version of possibility theory, launched by one English economist Shackle [39], uses a pair of dual set-functions (possibility and necessity measures) instead of only one to help capture partial ignorance [11]. Besides, it is not additive but highlights the key axiom of "maxitivity", and makes sense on ordinal structures that directs to a branch of the qualitative possibility theory [11]. Among various applications of possibility theory [4,8,12,19,23,31,34,36,37,38] that we cited, refs. [4,38] belong to the framework of Shackle's version. Refs. [23,31,34,36,37] fall into the version that follows Zadeh's view of relating possibility with membership of fuzzy sets [44], which highlights the key axiom of "maxitivity" and is appropriate for handling fuzzy uncertainty that arose in, e.g., natural language. Refs. [12,19] regard possibility as degree of membership but without resorting to the maxitivity axiom. Ref. [8] treats possibility as subjective probability. Our work only focusses on the version of Zadeh, who coined the name Theory of Possibility that highlights the key axiom of "maxitivity". Recall that probability is additive in disjunctive operation of mutually-exclusive events. We hence denote probability theory and Zadeh's version of possibility theory hereafter as sigma system and max system, respectively [32]. Following the line of Zadeh, possibility theory is gradually growing to exhibit itself as a potential foundation for fuzzy sets [31]. Membership function of fuzzy sets was recognized in [9] as likelihood function of possibility (instead of regarding membership function as possibility by Zadeh), and composition of fuzzy relations was found to be equal to composition of conditional possibilities [31].

The major jobs for achieving the objective of this paper will be elaborated in the following sections. In Section 2, we make analysis on the process of concept cognition and present mathematical definitions for intension/extension of concept, and introduce the subsethood measure for characterizing the confidence of concept classification. In Section 3, we present mathematical definitions of randomness and fuzziness, where we will see the occurring of fuzziness is closely related to the intension of the concept being classified. The subsethood measure, as a metric of fuzziness, is used in Section 4 for intuitive definition of possibility. In Section 5, the axiomatic definitions of probability and possibility are abstracted from their intuitive definitions by taking advantage of properties of randomness and fuzziness, respectively. We derived the conclusion that "max" is the only but un-strict disjunctive operator that is applicable across the fuzzy event space, and is an exact operator for fuzzy feature extraction, i.e., for extracting the value from the fuzzy sample space that leads to the largest possibility of one. In Section 6, the induced sigma-max system is presented along with a unified statistical interpretation, and the principle for the choice of sigma/max operators is analyzed. In Section 7, an example of stock price prediction using real dataset is given to demonstrate the sigma-max system.

## 2. The origin of fuzziness: concept cognition and subsethood measure

The origin of fuzziness is closely related to the process of concept cognition, which is to be analyzed by a fusion of the achievements from both the area of logics and cognition psychology and the area of artificial neural network (ANN). We reach the conclusion that a concept is determined is equivalent to either its intension or extension is determined, which is the same as that of the first order predicate logic (FOPL) [10].

### *2.1 Concept and its intension/extension*

According to Collins Discovery Encyclopedia, a concept is a general idea or notion that corresponds to some class of entities and that consists of the characteristic or essential features of the class [47]. In Merriam-Webster.com Dictionary, a concept is defined as an abstract or generic idea generalized from particular instances [48]. Concepts are "nothing but abbreviations in which we comprehend a great many different sensuously perceptible things according to their common properties" [52].

A concept may be analyzed into its intension (content) and extension (range). According to Encyclopedia Britannica, intension indicates the internal content (features or properties) of a concept that constitutes its formal definition; and extension, as counterpart of intension, indicates its range of applicability by naming the particular objects that it denotes [49]. According to [47], the intension of a concept refers to the totality of essential properties by reference to which the objects in a given concept are generalized and differentiated; the extension refers to the totality of generalized objects reflected in the concept. For instance, the intension of the concept "automobile" as a substantive is "vehicle driven by engine for conveyance on road," whereas its extension embraces such things as cars, trucks, buses, and vans.

It is widely acknowledged that the correlative intension/extension follow the law of inverse relation [47,49]. That is the wider the set of properties (i.e., the more restricted the intension), the narrower the extension of the concept, and vice versa. The intersection of an extension corresponds logically to the union of the intension [17,49].

Based on the above introduction, definitions of concept, intension and extension are selected and summarized below. Some remarks are then followed to show our perspectives, some of which are based on discussions of concept cognition in the next subsection.

**Definition 2.1.** A *concept* is defined as an abstract idea or pattern generalized from particular instances or objects.

**Definition 2.2.** *Intension* of a concept indicates the abstracted internal feature of a concept that constitutes its formal definition.

**Definition 2.3.** *Extension* of a concept refers to the totality of objects reflected in the concept.

*Remarks:*

1) It is the abstracted feature (intension) that defines or is equivalent to a certain concept. The abstracted feature, as will be illustrated in Example 2.1 and Fig. 1a below, is a kind of feature extracted by human cognition from the natural selected feature.

2) The intension of a concept, as will later be discussed in Section 2.3, could be modeled as either a vector or a set. The set of intensions can be simply denoted by an ellipse $f$ no matter it is extracted from one feature or multiple features. For a class of concepts, their intensions could be modeled as a family of ellipses, each of which refers to the intension of a certain concept.

3) Intension and extension should be stated to follow a law of proportional relation, instead of the law of inverse relation. As discussed above, the more restricted the intension of a concept, the narrower its extension. To our viewpoint, more restricted means the domain of the intension (ellipse) is smaller.

4) By the FOPL, a concept is determined is equivalent to either the intension or the extension of a concept is determined [10]. The smaller (i.e., more restricted) the intension, the narrower the extension (scope) of the concept, and vice versa.

5) A great number of concepts are fuzzy concepts, the extensions of which are blurring or not clear, which is due to the overlapping of the intensions of different concepts. The definition of fuzzy concept will be given later in Definition 3.10.

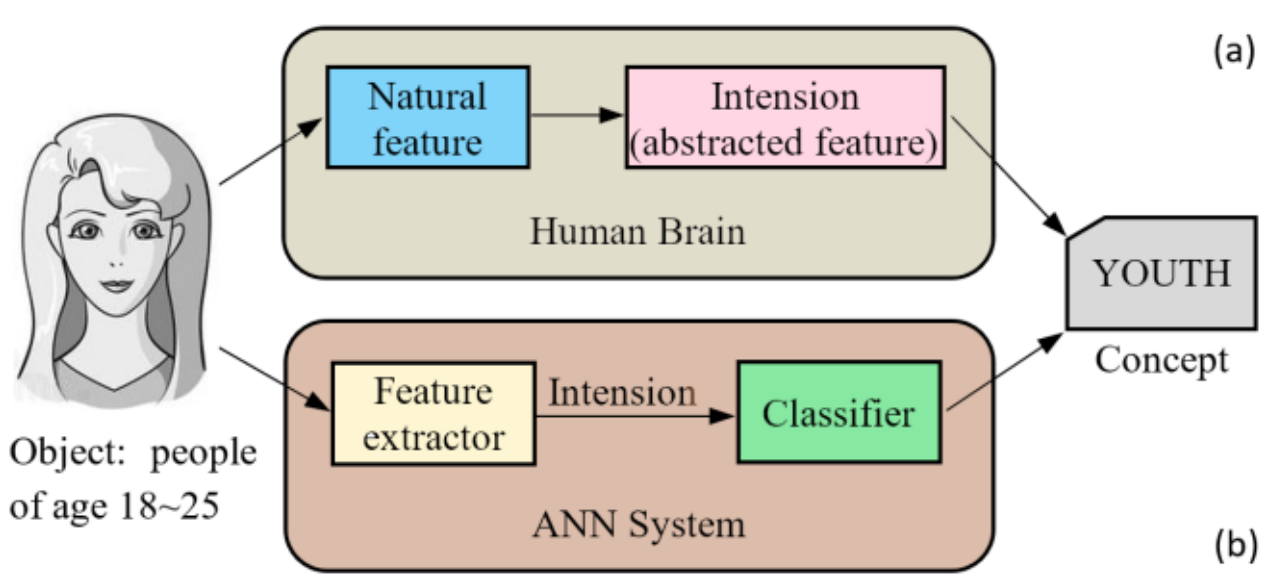

Fig. 1. Concept cognition. (a) The process of concept cognition in human brain. (b) The structure of an ANN system.

**Example 2.1.** Age can be a natural and selected feature variable for defining the concept of YOUTH. In general, you would regard people with age of 18 or 25 as YOUTH whereas regard age of 50 not as YOUTH. Obviously, ages of 18 and 25 indicate different values for the feature variable of age. Here comes a question, why do you regard age of 18 or 25 as YOUTH? The deeper reason, as shown in Fig. 1a, is that there should be an abstracted feature that you formed in your mind according to the age, by which you make your judgement and build up the concept of YOUTH. In this case, age of 18 and 25 delivered equivalent information of intension for the concept of YOUTH. Though you may feel hard to tell concretely what the abstracted feature is, it must be there. On the other hand, it is possible that the abstracted feature could be learned by an ANN classifier as will be introduced in the next subsection. Similarly, we can judge whether a person is a youth or not by his appearance. Here appearance is also a natural and selected feature, from which we human being can form an abstracted feature, as well, by which we can make a judgement whether this person is a youth or not.

### *2.2 An ANN perspective of concept cognition*

We in this subsection use an ANN system to illustrate the process of concept cognition in human brain and help model the terms of intension and extension. By concept cognition we mean concept learning and concept classification. Modern deep ANN systems, e.g., the convolutional neural network (CNN), can well imitate the structure and function of human brain [21,29]. A deep ANN system $s$ may have many layers, but can generally, as shown in Fig. 1b, be divided into two subfunctions such as feature extractor $s_1(\cdot)$ and classifier $s_2(\cdot)$.

By the training procedure, system $s$ can learn to build up some patterns or concepts labeled $\{x_j\}$ $(j = 1,2,\ldots,\text{n})$ from a group of object samples $\{z_i\}$ $(i = 1,2,\ldots,\text{m})$, which refer to feature data of the object. The ANN system $s$ is said to be well trained when predetermined requirements are achieved. Knowledge of these learned concepts $\{x_j\}$ is then represented by the function $s_2(s_1(\cdot))$ of the system $s$, which fulfills the mapping from input samples $\{z_i\}$ into output concepts $\{x_j\}$. Denote all input samples that corresponding to label $x_j$ as a single equivalent sample $y_j$, then the output $f_{x_j}$ of feature extraction is

$$f_{x_j} = s_1(y_j) \,. \qquad (1)$$

We call $f_{x_j}$ the intension of concept $x_j$, which is generated from the equivalent sample $y_j$ by the function $s_1(\cdot)$ and points to label $x_j$ through the classifier $s_2(\cdot)$, i.e.,

$$x_j = s_2(f_{x_j}) = s_2(s_1(y_j)) \,. \qquad (2)$$

A well-trained system $s$ can be deployed for practical work of classification. For example, given an object sample $z_t$, we would like to know which label $X$ the system $s$ would classify it into, i.e.,

$$X = s_2(f_X) = s_2(s_1(z_t)), \qquad (3)$$

where intension $f_X$ is the output of feature extraction given sample $z_t$, and $X \in \{x_j\}$ $(j = 1,2,\ldots,\text{n})$.

By collecting all samples labeled $x_j$, denoted by

$$e_{x_j} = \{z|s_2(s_1(z)) = x_j\}, \qquad (4)$$

we come to recognize the extension $e_{x_j}$ of the concept $x_j$, which indicates the scope of the objects that concept $x_j$ can apply to. Therefore, a concept can be mathematically described as in Postulate 2.1; whereas an object can be modeled by a triplet of $(z, x_j, f_{x_j})$ or $(z, X, f_X)$, where $z$ is sample, $x_j$ and $X$ are label or unknown label, and $f_{x_j}$ and $f_X$ are intension.

**Postulate 2.1.** A concept can be formed by a concept learner as either human brain or the ANN-based, and mathematically modeled by a triplet of $(x_j, f_{x_j}, e_{x_j})$, where $x_j$ is label, $f_{x_j}$ is the intension as modeled by (1), and $e_{x_j}$ is the extension as modeled by (4).

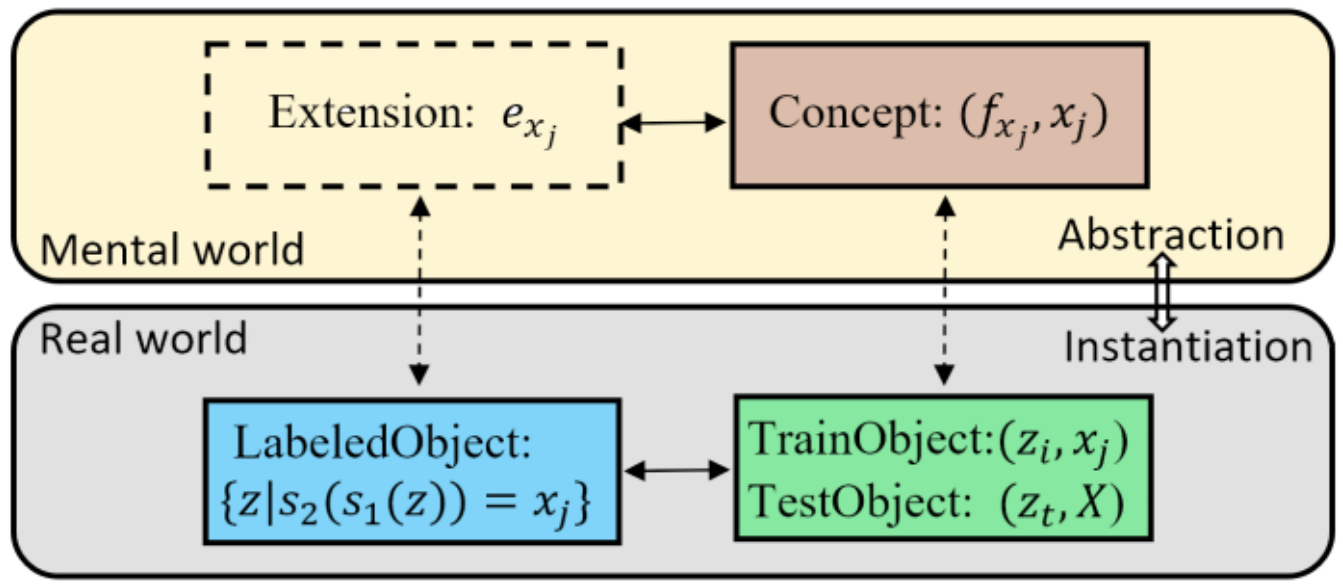

Fig. 2. A two-world model of concept cognition.

The process of concept cognition can as well be visualized by a two-world model as Fig. 2, where concept and its intension/ extension belong in the mental world of human being, which are built up from objects of the real world. We by comparing their intensions $f_X$ and $f_{x_j}$ determine whether a given object with unknown label $X$ (object $X$ in short) belongs to a certain concept $x_j$. By the process of concept cognition analyzed in this subsection, we reach the same conclusion as the FOPL that a concept is determined is equivalent to either its intension or extension is determined.

### *2.3 The measure of fuzziness: subsethood measure*

For the ANN system, intension $f_{x_j}$ is usually in form of feature vector or feature map [29,42]. A feature map may consist in a set of binary features, on which set operations ( like ∩ and ∪) can be defined. The equivalence between the forms of vector and set can be further visualized by Fig.3. Let $c = (x_1, y_1)$ and $d = (x_2, y_2)$ be two feature vectors, which can then be represented by two arrows ($\overrightarrow{oc}$ and $\overrightarrow{od}$) and equivalently by the determined two rectangles in yellow and blue. Since we are familiar with Venn Diagram, then set operations can be defined on the two rectangles. In what followed, we will simply use ellipse to visualize the set form of features.

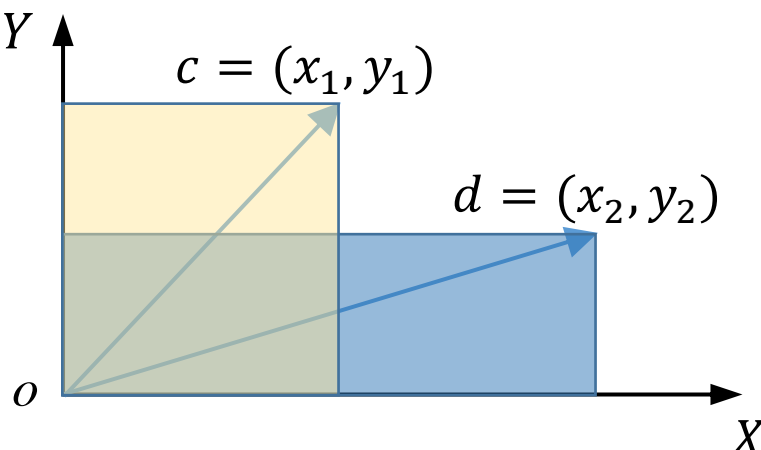

Fig. 3. Equivalence between forms of vector and set.

Most ANN systems use similarity (or distance) measure to classify objects [30,42], where the tested object will be assigned to the category that has the largest similarity measure. For example, cosine distance as below was used in [30] to classify the shape of an object, where $f_X$ and $f_{x_j}$ are given in forms of feature vectors.

$$\cos(f_X, f_{x_j}) = \frac{|f_X \cdot f_{x_j}|}{|f_X||f_{x_j}|} \qquad (5)$$

where "·" is the inner product and $|\cdot|$ is the L-2 norm. The cosine distance of (5) is comparable to the set-based similarity measure below [24,40] when $f_X$ and $f_{x_j}$ are in forms of sets.

$$\text{sim}(f_X, f_{x_j}) = \frac{|f_X \cap f_{x_j}|}{|f_X \cup f_{x_j}|} \qquad (6)$$

where $|\cdot|$ denotes the cardinal number of sets, and "sim" means similarity. This work would suggest to denote $|\cdot|$ as size instead of cardinal number, where size indicates "cardinality" for countable set or "measure" for real number set $\mathbb{R}^n$.

The analogy of (5) and (6) could be spotted by two typical cases below:

1) $\cos(f_X, f_{x_j}) = 1$ when vectors $f_X$ and $f_{x_j}$ are parallel ⟷ $\text{sim}(f_X, f_{x_j}) = 1$ when sets $f_X = f_{x_j}$.

2) $\cos(f_X, f_{x_j}) = 0$ when vectors $f_X$ and $f_{x_j}$ are vertical ⟷ $\text{sim}(f_X, f_{x_j}) = 0$ when sets $f_X \cap f_{x_j} = \emptyset$.

Considering that an object $X$ could be classified into a concept labeled $x_j$ if their intensions, in forms of sets, satisfy $f_X \subseteq f_{x_j}$ (instead of $f_X = f_{x_j}$), we decide to use subsethood measure $\text{degree}(f_X \subseteq f_{x_j})$, closely related to the similarity measure, to characterize the confidence of concept classification, $\text{degree}(X = x_j)$, which is defined as [24,40]

$$\text{degree}(X = x_j) = \text{degree}(f_X \subseteq f_{x_j}) = \frac{|f_X \cap f_{x_j}|}{|f_X|} \qquad (7)$$

where $|\cdot|$ denotes the cardinality or measure of sets.

If we rewrite (5) as below, then it can be easily observed that (5) and (7) are comparable, as well.

$$\cos(f_X, f_{x_j}) = \left| f_X \cdot \frac{f_{x_j}}{|f_{x_j}|} \right| / |f_X| \qquad (8)$$

where "·" is the inner product and $|\cdot|$ is the L-2 norm. The subsethood measure of (7), as a metric of fuzziness, will later be used for definitions of fuzziness and possibility.

To end this section, we would like to continue the discussion of Example 2.1. Note that the feature extractor as shown in Fig. 1b can be seen as distorting the input in a non-linear way so that categories become linearly separable by the last layer of classifier [29]. Therefore, by feeding samples, e.g., of age varying from 7 to 45 with labels of JUVENILE, YOUTH and MID-LIFE (MID), into an ANN concept learner, it is possible that the intension of YOUTH could be learned and visualized. It could be imagined that ages of 18 and 25 will be mapped into the same point in the feature map whereas ages of 28 and 35 would most likely be projected into significantly different places in the feature map because of the non-linear function of the feature extractor.

## 3. The definitions of randomness and fuzziness

Human cognition of randomness and fuzziness has a long history, discussions of them can be found from time to time [22,25,35,41,45], yet clear and widely-accepted definitions for them remain elusive. Randomness and fuzziness are obviously different but very easy to be mixed up. Nearly all books on probability/possibility give definitions to probability/possibility, but none give definitions to randomness/fuzziness. We below give their definitions and the related in forms of both text description and mathematic formulation. The text part of Definitions 3.1 and 3.5 is adapted from [33] with a slight refinement.

**Definition 3.1.** *Randomness* is the occurrence uncertainty of the either-or outcome of a causal experiment, characterized by the lack of predictability in mutually exclusive outcomes [33]. Mathematically, given the causal experiment *X* with *N* possible outcomes $\Omega = \{x_1, x_2, \ldots\ldots, x_N\}$ that satisfies $\forall x_i \neq x_j$ we have $\{X = x_i\} \cap \{X = x_j\} = \emptyset$, the randomness of the uncertain experiment could be characterized by: $\exists! x_i$ so that $\{X = x_i\}$.

*Remark*: The occurrence uncertainty of the either-or outcome of a causal experiment can be easily interpreted as: the uncertainty of generating one and only one results from among multiple possible outcomes in a causal experiment. $\exists! x_i$ means there exists one and only one $x_i$. A causal experiment is any procedure of observing an uncertainty state *X* that can be infinitely repeated and has a well-defined set of possible outcomes $\Omega$. For simplicity, we can use experiment *X* to represent the experiment of observing the uncertainty state *X*. Mutually exclusive outcomes are mathematically modeled by

$$\{X = x_i\} \cap \{X = x_j\} = \emptyset, \forall x_i \neq x_j. \qquad (9)$$

Dice-tossing is a classic example of illustrating random experiment or randomness. Obviously, each time one can only get one side among all six possible sides, which is unpredictable and is either this or that (e.g., either two or three). In brief, randomness is the occurrence uncertainty of the either-or outcome of a causal experiment.

**Definition 3.2.** *Random sample space* $\Omega$ is the set of the elementary random events, which consists of all possible mutually-exclusive outcomes of a causal experiment. Mathematically, random sample space $\Omega = \{x_1, x_2, \ldots\ldots, x_N\}$, and $\forall x_i \neq x_j$ we have $\{X = x_i\} \cap \{X = x_j\} = \emptyset$.

**Definition 3.3.** *Random event space* is the $\sigma$-algebra $F \subseteq 2^{\Omega}$, which consists of a set of events $\{A_i\}$.

*Remark*: Each event $A_i$ is a set containing zero or groups of outcomes which might be of more practical use, whereas an outcome is the result of a single execution of the experiment. $2^{\Omega}$ is the power set of $\Omega$, which consists of all subsets $A_i s$ of $\Omega$. $F$ is a $\sigma$-algebra, which means $F$ is closed under operations of complement, countable union and intersection. Random events in event space $F$ may no longer be mutually exclusive. However, the natural attribute of randomness, as claimed in Definition 3.1, is mutually exclusive since the elementary random events in the random sample space $\Omega$ are mutually exclusive. It should also be noted that probability is axiomatically defined upon mutually exclusive events.

**Definition 3.4.** A *random variable X* is a variable whose value $x_i$ is subject to variations due to random uncertainty.

*Remark*: A random variable can take on a set of possible values in a random sample space $\Omega$, or its generated event space $F \subseteq 2^{\Omega}$. By introducing random variable *X* (or fuzzy variable below), we in this work mix the use of $x_i$ and $A_i$.

**Definition 3.5.** *Fuzziness* is the classification uncertainty of the both-and outcome of a cognition experiment, characterized by the lack of clear boundary between non-exclusive outcomes [33]. Mathematically, given the cognition experiment *X* with *N* possible outcomes $\Psi = \{x_1, x_2, \ldots\ldots, x_N\}$ that satisfies $\exists x_i, x_j$ so that $f_{x_i} \cap f_{x_j} \neq \emptyset$, the fuzziness of the uncertain experiment could be characterized by: $\exists x_i, x_j$ so that $f_X \cap f_{x_i} \cap f_{x_j} \neq \emptyset$.

*Remark*: The classification uncertainty of the both-and outcome of a cognition experiment can be easily interpreted as: the uncertainty that occurs in simultaneously classifying an object into multiple possible concepts with associated confidences in a cognition experiment. Concepts $x_i$ and $x_j$ are said to be non-exclusive if their intensions $f_{x_i}$ and $f_{x_j}$ satisfy

$$f_{x_i} \cap f_{x_j} \neq \emptyset. \qquad (10)$$

As shown in Fig. 4a, $\exists x_i, x_j$ so that

$$f_X \cap f_{x_i} \cap f_{x_j} \neq \emptyset \qquad (11)$$

is the mathematical formulation for describing "simultaneously classifying an object into multiple possible concepts with associated confidences". Note that by $f_X \cap f_{x_i} \cap f_{x_j} \neq \emptyset$ we can derive $0 < \text{degree}(X = x_i) \leq 1$ and $0 < \text{degree}(X = x_j) \leq 1$, as given in Proposition 3.1. A cognition experiment is any procedure of classifying an object *X* that can be infinitely repeated and has a well-defined set of possible outcomes $\Psi$. Both the causal/random experiment and the cognition/fuzzy experiment could be repeated experimentation. That means the sigma/max inferences to be induced from randomness and fuzziness could both be statistical, see Section 6.3 for a unified statistical interpretation of sigma-max system. Similarly, we use experiment *X* to concisely represent the experiment of classifying an object *X*. The classification of age-group is a classic example of illustrating fuzzy experiment or fuzziness. Suppose you are informed of a person of 40 years old and you are invited to classify this person into age-group = {YOUTH, MID and AGED}, which is a procedure of classifying an object into multiple possible concepts. Most likely you will not consider years of 40 as AGED, but would regard years of 40 as both YOUTH and MID with associated confidences of 0.5 and 1, respectively. The outcomes of the classifying experiment are non-exclusive (both YOUTH and MID) and uncertain (with associated confidences), due to the lack of clear boundary between (the intensions of) them. In brief, fuzziness is the classification uncertainty of the both-and outcome of a cognition experiment.

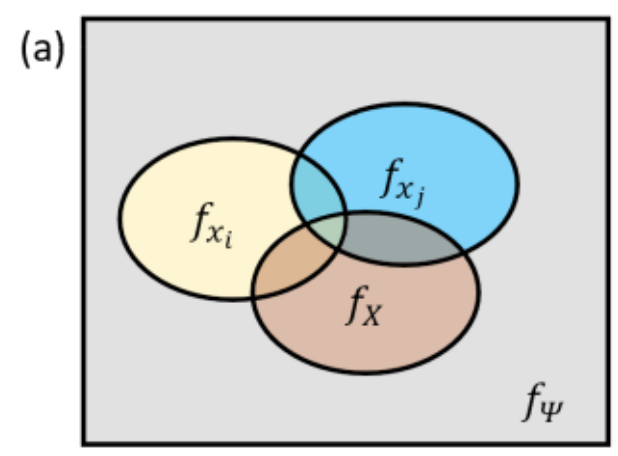

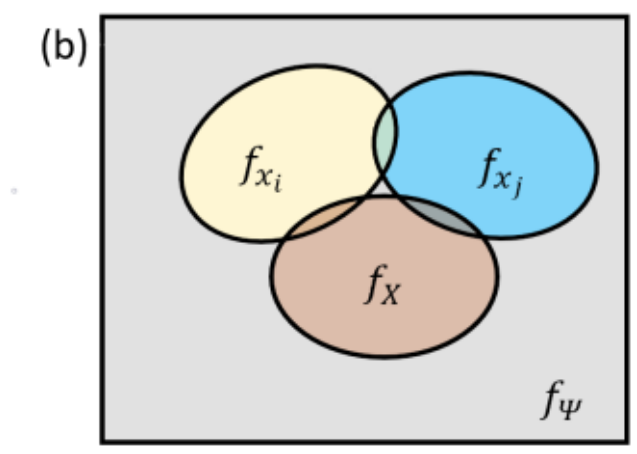

Fig. 4. (a) Projection-nonexclusive: $f_X \cap f_{x_i} \cap f_{x_j} \neq \emptyset$. (b) Projection-exclusive: $f_{x_i} \cap f_{x_j} \neq \emptyset$ & $f_X \cap f_{x_i} \cap f_{x_j} = \emptyset$.

**Proposition 3.1.** If $\exists x_i, x_j$ so that $f_X \cap f_{x_i} \cap f_{x_j} \neq \emptyset$, then we have $0 < \text{degree}(X = x_i) \leq 1$ and $0 < \text{degree}(X = x_j) \leq 1$.

*Proof*: From $f_X \cap f_{x_i} \cap f_{x_j} \neq \emptyset$, we have $f_X \cap f_{x_i} \neq \emptyset$ and $f_X \cap f_{x_j} \neq \emptyset$. By (7), we have $0 < \text{degree}(X = x_i) \leq 1$ and $0 < \text{degree}(X = x_j) \leq 1$. Hence Proposition 3.1 holds.

**Definition 3.6.** *Fuzzy sample space* $\Psi$ is the set of the elementary fuzzy events, which consists of all possible non-exclusive outcomes of a cognition experiment. Mathematically, fuzzy sample space $\Psi = \{x_1, x_2, \ldots\ldots, x_N\}$, and $\exists x_i$,$x_j$ so that $f_X \cap f_{x_i} \cap f_{x_j} \neq \emptyset$.

*Remark*: To be a fuzzy sample space, the condition $f_{x_i} \cap f_{x_j} \neq \emptyset$ is not sufficient since if $\forall x_i \neq x_j$ we have $f_X \cap f_{x_i} \cap f_{x_j} = \emptyset$, then as shown in Fig. 4b such a fuzzy sample space would be trivial and useless. The case $f_X \cap f_{x_i} \cap f_{x_j} = \emptyset$ is called projection-exclusive, i.e., intensions of $x_i$ and $x_j$ are mutually exclusive when projected into the intension of *X*.

**Definition 3.7.** *Exhaustive fuzzy sample space* $\Psi^+$ is the fuzzy sample space $\Psi$ that satisfies $\exists x_i$ so that $\text{degree}(X = x_i) = 1$.

*Remark*: In this paper, it is assumed that fuzzy sample space $\Psi$ is always exhaustive unless it is specifically pointed out. To simplify the notion, we will use fuzzy sample space $\Psi$ to loosely refer to exhaustive fuzzy sample space $\Psi^+$.

**Definition 3.8.** *Fuzzy event space* is the $\sigma$-algebra $\Sigma \subseteq 2^\Psi$, which consists of a set of events $\{A_i\}$.

*Remark*: It is in the sense of their intensions ($f_{x_i} \cap f_{x_j} \neq \emptyset$) that elements $x_i$ and $x_j$ of $\Psi$ are non-exclusive. Fuzzy sample space $\Psi$ is a classic set instead of a fuzzy set, and fuzzy event space $\Sigma \subseteq 2^\Psi$ is a Boolean algebra and then a $\sigma$-algebra. Though events in $F \subseteq 2^\Omega$ and $\Sigma \subseteq 2^\Psi$ are both not mutually exclusive, the structures of random event space $F \subseteq 2^\Omega$ and fuzzy event space $\Sigma \subseteq 2^\Psi$ are not the same because their corresponding sample spaces $\Omega$ and $\Psi$ are defined differently. This point will be further revealed later in the analysis of disjunctive operations of probability and possibility.

**Definition 3.9.** A *fuzzy variable X* is a variable whose value $x_i$ is subject to variations due to fuzzy uncertainty.

*Remark*: A fuzzy variable can take on a set of possible values in a fuzzy sample space $\Psi$, or its generated event space $\Sigma \subseteq 2^\Psi$.

**Definition 3.10.** *Fuzzy concept* (*event*) is the element of the fuzzy event space.

## 4. Intuitive definitions of probability and possibility

The intuitive definitions of probability and possibility are defined on the random sample space $\Omega$ and the fuzzy sample space $\Psi$, respectively.

### *4.1 Intuitive definition of probability*

**Definition 4.1.** *Probability* (classic frequency definition) $p_X(x_i)$ is the measure of the either-or randomness, which is the frequency of generating outcome $x_i$ in an experiment of observing the uncertainty state *X*. Probability $p_X(x_i)$ can be numerically described by the vote number $n_i$ of desired outcomes $x_i$ divided by the total votes $n_t$ of all outcomes, as calculated by:

$$p_X(x_i) = \lim_{n_t \to \infty} \frac{n_i}{n_t} \quad (\sum n_i = n_t). \tag{12}$$

*Remark*: Note that $p_X(x_i)$ is usually and hereafter denoted as $p(x_i)$ for short. Given $\Omega = \{x_1, x_2, \ldots, x_N\}$, it is straightforward to have

$$p(\cup_{i=1}^N x_i) = \frac{n_1 + n_2 + \cdots + n_N}{n_t} = \Sigma_{i=1}^N p(x_i) = 1. \tag{13}$$

As an extreme case when all votes of $\Omega$ fall into $x_i$, i.e., $n_i = n_t$, we have

$$p(x_i) = \frac{n_i}{n_t} = 1. \tag{14}$$

The case that $p(x_i) = 1$ can exist for only one of the elementary events among the random sample space $\Omega$, i.e., distinct values of $x_i$ cannot simultaneously have a degree of probability equal to 1.

*4.2 Intuitive definition of possibility*

An intuitive definition of possibility based on compatibility interpretation was given in [33]. However, the concept of "compatibility" itself needs to be defined mathematically with a deeper physical explanation. Definition 4.2 below refines the existing definition of intuitive possibility.

**Definition 4.2.** *Possibility* (intuitive definition) $\pi_X(x_i)$ is the measure of the both-and fuzziness, which is the confidence of classifying an object $X$ into concept $x_i$. Possibility $\pi_X(x_i)$ of the outcome $x_i$ can be numerically described by the compatibility between a fuzzy variable $X$ and its prospective outcome $x_i$, which can be defined as

$$\pi_X(x_i) = \mathrm{comp}(X, x_i) = \mathrm{degree}(f_X \subseteq f_{x_i}) = \frac{|f_X \cap f_{x_i}|}{|f_X|} \quad (15)$$

where "comp" means compatibility; $f_X$ and $f_{x_i}$ are sets of intensions of the fuzzy variable $X$ and the concept $x_i$, respectively; and $|\cdot|$ is the cardinality or measure.

*Remark*: The determination of $f_X$ and $f_{x_i}$ in Definition 4.2 can be achieved through various statistical experiments, such as the training process of deep learning as exemplified by (1) ~ (3). In certain instances, it may suffice to ascertain $|f_X \cap f_{x_i}|$ and $|f_X|$. An example of this will be provided in Section 6.3. Also note that $\pi_X(x_i)$ is usually and hereafter denoted as $\pi(x_i)$ for short. Compatibility may be replaced by another better word "membership". In Zadeh's fuzzy set theory, membership is usually denoted as $\mu_F(x_i)$ to indicate the compatibility of $x_i$ with the concept labeled $F$. An improved denotation $\mu_{Y|X}(F|x_i)$ was suggested for membership in [31], to indicate the compatibility of fuzzy concept variable $Y$ with the concept labeled $F$ given its fuzzy attribute variable $X$ being $x_i$. Mathematically, Zadeh's membership equals to conditional possibility whereas membership function equals to likelihood function of possibility [9,31]. In our work, membership is denoted as $\mu_X(x_i)$, to indicate the degree of applicability of a concept $x_i$ to a fuzzy variable $X$ given null condition. In this view, $\mu_X(x_i)$ and $\pi_X(x_i)$ are equivalent.

*4.3 Discussions on intuitive possibility*

Eq. (15) is to be further discussed and explained with examples below.

1) Suppose, as shown in Fig. 5a, the intension of fuzzy variable $X$ falls into the intension of a special concept of the sample space $\Psi$, i.e., $f_X \subseteq f_\Psi$, then we have

$$\pi(\Psi) = \frac{|f_X \cap f_\Psi|}{|f_X|} = \frac{|f_X|}{|f_X|} = 1. \quad (16)$$

**Example 4.1.** An example of Fig. 5a may be that a person of age around 40 can always be considered as a human being. Here a person of age around 40 (be categorizing) can be taken as the fuzzy variable $X$, and "human being" is a special concept $\Psi$.

2) Suppose, as shown in Fig. 5b, the intersection of the intensions of fuzzy variable $X$ and the concept $x_i$ is not empty, i.e., $f_X \cap f_{x_i} \neq \emptyset$, then we have

$$0 < \pi(x_i) = \frac{|f_X \cap f_{x_i}|}{|f_X|} \leq 1. \quad (17)$$

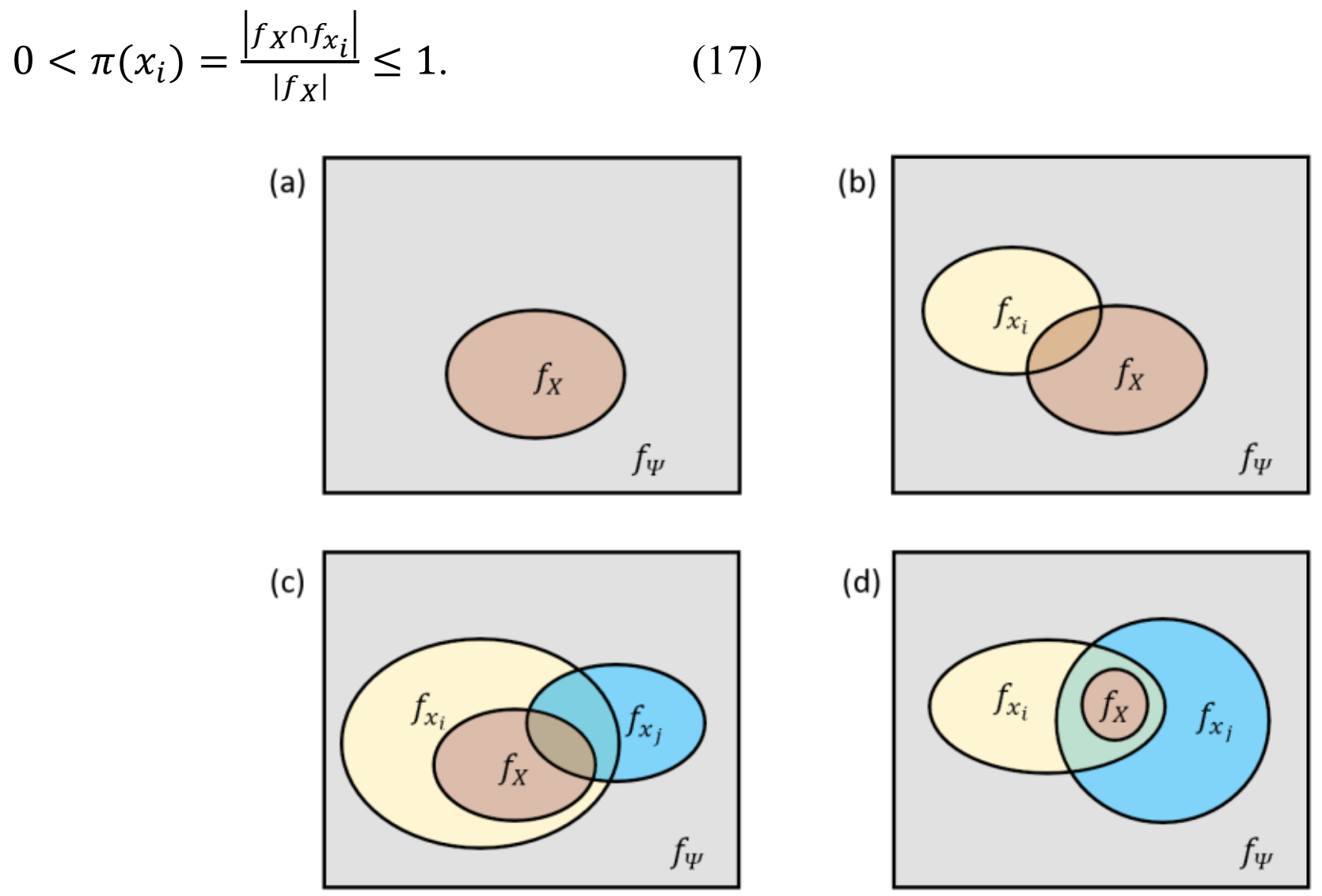

Fig. 5. (a) $f_X \subseteq f_\Psi$. (b) $f_X \cap f_{x_i} \neq \emptyset$. (c) $X$ is exhaustive: $f_X \subseteq f_{x_i}$. (d) $X$ is innocent: $\pi(x_i) = \pi(x_j) = 1$.

**Example 4.2.** Suppose you are walking on the street with a friend of yours and you are asked to what extent you would rate your friend as a tall person. You definitely have some knowledge in your mind about the height of your friend and have your standard of categorizing a person's height, which means you know the intension $f_X$ of the fuzzy variable (relating to your friend's height) and the intension $f_{x_i}$ of the concept "tall person". Then you can either directly assign a value between 0 and 1 or figure out a value by using (15) to represent your rating. In the latter case, you will need to quantify $f_X$ and $f_{x_i}$ according to your knowledge in your mind. Obviously, the quantification of $f_X$ and $f_{x_i}$ by you is subjective and rather arbitrary, which nevertheless can be learned by a deep ANN classifier if samples related to "height" are available.

3) Normalized possibility and sub-normalized possibility. If the intuitive possibility is defined on the exhaustive fuzzy sample space $\Psi^+$, then it is normalized possibility. In other words, for the normalized possibility, it would be a mandatory demand that fuzzy variable *X* is exhaustive. That is at least one of the elements of *X* should be the actual world, mathematically, $\exists x_i$ such that $f_X \subseteq f_{x_i}$ as shown in Fig. 5c, and we have

$$\pi(x_i) = \frac{|f_X \cap f_{x_i}|}{|f_X|} = 1 \tag{18}$$

If the intuitive possibility is defined on the fuzzy sample space $\Psi$ and $\forall x_i$ we have $\pi(x_i) < 1$, then it is sub-normalized possibility. In this paper, it is assumed that possibility is always normalized unless it is specifically pointed out.

Since possible values of fuzzy variable *X* among $\Psi$ are non-exclusive, we can simultaneously have $\pi(x_i) = 1$ for any numbers of elementary events among the fuzzy sample space $\Psi$. That should be one of the major distinctions of possibility from probability, recall that $p(x_i) = 1$ can only exist for a certain elementary event $x_i$ among the random sample space $\Omega$. Given $\Psi = \{x_1, x_2, ..., x_N\}$, we generally have

$$\Sigma_{i=1}^{N} \pi(x_i) \neq 1 \tag{19}$$

which is also different from (13) of probability.

4) Representation of innocent. That means fuzzy variable *X* can take all possible values with possibility of one, i.e., $\pi(x_i) = \pi(x_j) = 1$ as shown in Fig. 5d.

Overall, Eq. (15) provides a mathematical formulation with physical notion for the definition of possibility, which would promote the theoretical exploration of the possibility theory. In the next section, we will introduce the abstraction of the axiomatic definition of possibility by the application of (15).

## 5. The abstraction of axiomatic probability/ possibility definitions

Probability and possibility are axiomatically defined on the event spaces $F \subseteq 2^{\Omega}$ and $\Sigma \subseteq 2^{\Psi}$, respectively, on which disjunctive operations will be defined, as well. Be aware that we in this work use notation $x_i$ for elements of both sample and event spaces.

### *5.1 Disjunctive operation of probability*

Given events $x_i$ and $x_j$ as illustrated by Fig. 6, and their probabilities $p(x_i)$ and $p(x_j)$, we now need to figure out $p(x_i \cup x_j)$. From the intuitive definition of probability, we have

$$p(x_i \cup x_j) = \frac{n_{x_i \cup x_j}}{n_t} = \frac{n_i + n_j - n_{x_i \cap x_j}}{n_t}, \tag{20}$$

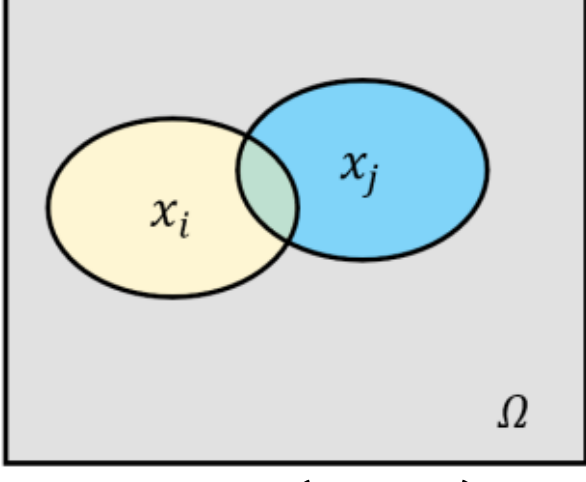

Fig. 6. $p(x_i \cup x_j)$.

It follows that

$$p(x_i \cup x_j) = p(x_i) + p(x_j) - p(x_i \cap x_j). \tag{21}$$

Therefore, we have

$$\max\{p(x_i), p(x_j)\} \leq p(x_i \cup x_j) \leq p(x_i) + p(x_j). \quad (22)$$

When $x_i$ and $x_j$ are mutually exclusive with no votes falling into $x_i \cap x_j$, i.e., $n_{x_i \cap x_j} = 0$, we have

$$p(x_i \cup x_j) = p(x_i) + p(x_j). \quad (23)$$

Note that derivation of (23) has once been discussed in [31] for the empirical deduction of the probability axioms.

When votes of $x_i$ and $x_j$ are nested with votes of either $x_i$ or $x_j$ always falling into $x_i \cap x_j$, i.e., $n_i = n_{x_i \cap x_j}$ or $n_j = n_{x_i \cap x_j}$, we have

$$p(x_i \cup x_j) = \max\{p(x_i), p(x_j)\}. \quad (24)$$

Be aware that the case that votes of $x_i$ and $x_j$ are nested is practically impossible when it is applied to the random sample space $\Omega$, and is trivial when it is applied to the generated random event space $F \subseteq 2^{\Omega}$, which would demand that $x_i \subseteq x_j$ or $x_j \subseteq x_i$. For example, suppose $\Omega = \{x_1, x_2, x_3\}$ and $F \subseteq 2^{\Omega} = \{\{x_1\}, \{x_2\}, \{x_3\}, \{x_1, x_2\}, \{x_1, x_3\}, \{x_2, x_3\}, \{x_1, x_2, x_3\}\}$. Then, applying of (24) to the random sample space $\Omega = \{x_1, x_2, x_3\}$ would demand that votes of $x_1$, $x_2$ and $x_3$ are nested, which violates the original setting of Definition 3.2, that they are mutually exclusive, i.e., $\forall x_i \neq x_j$ we have $\{X = x_i\} \cap \{X = x_j\} = \emptyset$. The max operation of (24) can only be applied to some, not all, elements of the random event space $F \subseteq 2^{\Omega}$, e.g., $\{x_1\}$, $\{x_1, x_3\}$, and $\{x_1, x_2, x_3\}$, upon which the max operation defined would be of no practical value if it cannot be applied to the random sample space $\Omega$.

We hence conclude that from the intuitive definition of probability, we managed to derive the nontrivial disjunctive operation of "sigma" as indicated by (23).

### *5.2 Axiomatic definition of probability*

**Definition 5.1.** *Probability* (axiomatic definition) was built up upon a probability space $(\Omega, F, P)$, which is a mathematical construct that models a real-world process consisting of events that occur randomly. A probability space consists of three parts [22,27]:

1) the random sample space $\Omega$ , as defined in Definition 3.2.
2) the σ-algebra $F \subseteq 2^{\Omega}$, which is the event space consisting of a set of events $\{x_i\}$.
3) the probability measure $P: F \longrightarrow [0,1]$, which is a function on $F$ such that it satisfies the three axioms below:

Axiom 1. (Nonnegativity Axiom) $p(x_i) \geqslant 0$ for any event $x_i$.
Axiom 2. (Normality Axiom) the measure of entire random sample space is equal to one: $p(\Omega) = 1$.
Axiom 3. (Additivity Axiom) for every countable sequence of mutually exclusive events $\{x_i\}$, we have

$$p(\cup_{i=1}^{\infty} x_i) = \Sigma_{i=1}^{\infty} p(x_i) \quad (25)$$

### *5.3 Disjunctive operation of possibility*

Given $\pi(x_i)$ and $\pi(x_j)$ as illustrated by Fig. 7a, we now need to figure out $\pi(x_i \cup x_j)$. According to (15) and with the ellipse assumption of the intension-sets, we have

(a) 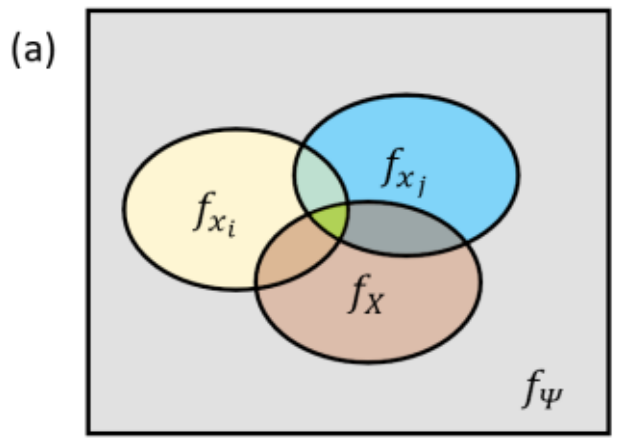

(b) 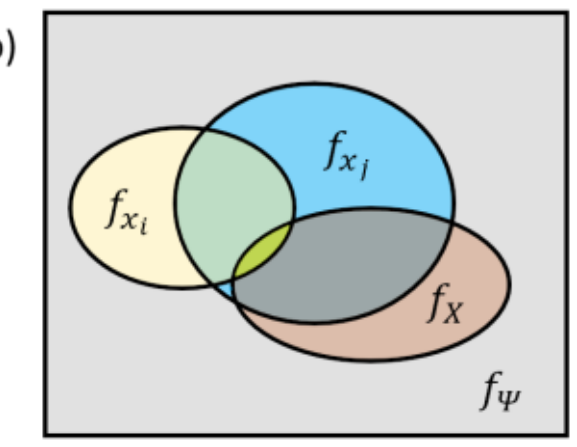

Fig. 7. (a) $f_X \cap \left(f_{x_i} \cup f_{x_j}\right)$. (b) Projection-nested ($f_X \cap f_{x_i} \subseteq f_X \cap f_{x_j}$).

$$\pi(x_i \cup x_j) = \frac{\left|f_X \cap f_{x_i \cup x_j}\right|}{|f_X|} = \frac{\left|f_X \cap f_{x_i} + f_X \cap f_{x_j} - f_X \cap f_{x_i} \cap f_{x_j}\right|}{|f_X|}$$
$$= \frac{\left|f_X \cap f_{x_i}\right| + \left|f_X \cap f_{x_j}\right| - \left|f_X \cap f_{x_i \cap x_j}\right|}{|f_X|} \quad (26)$$

where we assume

$$f_{x_i \cup x_j} = f_{x_i} \cup f_{x_j}\ , \qquad (27)$$

$$f_{x_i \cap x_j} = f_{x_i} \cap f_{x_j}\ , \qquad (28)$$

Assumptions (27) and (28) are reasonable considering that a concept is determined is equivalent to the intension of a concept is determined. From (26) it follows that

$$\pi\left(x_i \cup x_j\right) = \pi(x_i) + \pi\left(x_j\right) - \pi\left(x_i \cap x_j\right). \qquad (29)$$

Therefore, we have

$$\max\left\{\pi(x_i), \pi\left(x_j\right)\right\} \le \pi\left(x_i \cup x_j\right) \le \pi(x_i) + \pi\left(x_j\right) \quad (30)$$

When intensions of $x_i$ and $x_j$ are mutually projection-exclusive as previously shown in Fig. 4b, i.e., $(f_X \cap f_{x_j}) \cap (f_X \cap f_{x_i}) = \emptyset$, we have

$$\pi\left(x_i \cup x_j\right) = \pi(x_i) + \pi\left(x_j\right). \qquad (31)$$

The "sigma" operation as shown in (31) would become trivial once it is expanded into the fuzzy sample space as below

$$\pi(\cup_{i=1}^{N} x_i) = \Sigma_{i=1}^{N} \pi(x_i) = 1. \qquad (32)$$

Recall that for the normalized possibility, $\pi(x_i) = 1$ holds for at least one of the elements of *X*. Then (32) would demand that all other possibilities be zero. Even for sub-normalized possibility that satisfies $\forall x_i, \pi(x_i) < 1$, the "sigma" operation would potentially make $\pi(\cup_{i=1}^{N} x_i)$ calculated by (32) larger than one. Therefore, the "sigma" operation as shown in (31) is trivial and useless.

When intensions of $x_i$ and $x_j$ are projection-nested (nested when projected into the intension of the fuzzy variable *X*) as shown in Fig. 7b, i.e., $f_X \cap f_{x_i} \subseteq f_X \cap f_{x_j}$ or $f_X \cap f_{x_j} \subseteq f_X \cap f_{x_i}$, we have

$$\pi\left(x_i \cup x_j\right) = \max\left\{\pi(x_i), \pi\left(x_j\right)\right\} \qquad (33)$$

Note that *X* is exhaustive ($f_X \subseteq f_{x_j}$), as indicated by (18) and Fig. 5c, is a special case of being projection-nested. In the general case when intensions of $x_i$ and $x_j$ are neither projection-exclusive nor projection-nested and as indicated by (29) and Fig. 7a, the calculation of $\pi\left(x_i \cup x_j\right)$ will have to consider the value of $\pi\left(x_i \cap x_j\right)$ that is variable to $f_X \cap f_{x_i} \cap f_{x_j}$. In other words, in the general case there does not exist a disjunctive operator that is applicable across the fuzzy sample space (i.e., applicable to every pair of elements of the fuzzy sample space) for calculating $\pi\left(x_i \cup x_j\right)$ when given $\pi(x_i)$ and $\pi\left(x_j\right)$. We below present two more examples on the disjunctive operation of possibility, where fuzzy variable *X* may have more than two values.

**Example 5.1.** Fig. 8a shows us the case that situations of projection-exclusive and projection-nested both happened in classifying a given person of age 45 into age-groups, where three big ellipses respectively represent the intensions of YOUTH, MID and AGED, and one small ellipse indicates the intension of age 45. It is natural to assume that $f_{\text{age45}}$ fully falls into $f_{\text{MID}}$, and partially overlaps with $f_{\text{YOUTH}}$ and $f_{\text{AGED}}$, respectively. With this assumption, we see that $f_{\text{YOUTH}}$ and $f_{\text{MID}}$ are projection-nested, $f_{\text{AGED}}$ and $f_{\text{MID}}$ are projection-nested, and $f_{\text{YOUTH}}$ and $f_{\text{AGED}}$ are projection-exclusive. Therefore, we have

$$\pi(\text{YOUTH} \cup \text{MID}) = \max\{\pi(\text{YOUTH}), \pi(\text{MID})\} = \pi(\text{MID}),$$

$$\pi(\text{AGED} \cup \text{MID}) = \max\{\pi(\text{AGED}), \pi(\text{MID})\} = \pi(\text{MID}),$$

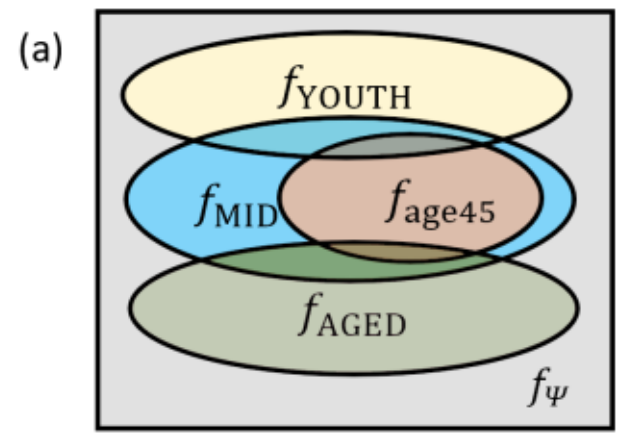

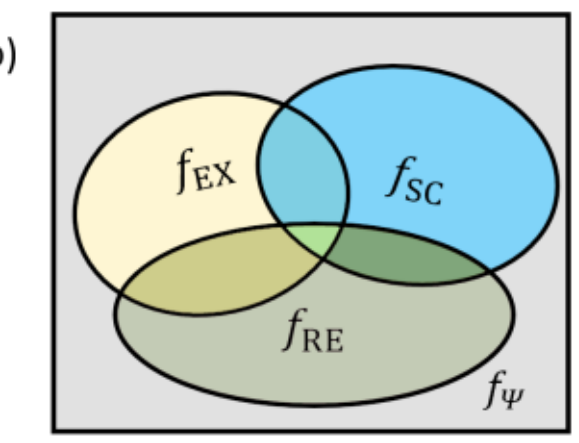

Fig. 8. (a) Projection-exclusive/nested. (b) Not projection-exclusive/nested.

$\pi(\text{YOUTH} \cup \text{AGED}) = \pi(\text{YOUTH}) + \pi(\text{AGED}).$

**Example 5.2.** Fig. 8b show us the case that a situation of neither projection-exclusive nor projection-nested happened in classifying a person of RESEARCHER (RE) into designated professions. Please be aware that the intension of the person being classified is not shown in Fig. 8b, which should be fully overlapped with that of RE. Obviously, the intension of RE is projection-nested with that of EXPERT (EX), and with that of SCHOLAR (SC), as well. And the intensions of EX and SC are neither projection-exclusive nor projection-nested. Therefore, we have

$$\pi(\text{EX} \cup \text{RE}) = \max\{\pi(\text{EX}), \pi(\text{RE})\} = \pi(\text{RE}),$$

$$\pi(\text{SC} \cup \text{RE}) = \max\{\pi(\text{SC}), \pi(\text{RE})\} = \pi(\text{RE}),$$

$$\pi(\text{EX} \cup \text{SC}) = \pi(\text{EX}) + \pi(\text{SC}) - \pi(\text{EX} \cap \text{SC}).$$

We hence conclude that from the intuitive definition of possibility, we managed to derive the disjunctive operation of "max" as indicated by (33) when the condition of being projection-nested is satisfied. In the next subsection, we will show how we relax the condition of being projection-nested, to lead to the well-known definition of possibility. By (21) and (29) and related analysis about the disjunctive operators of probability and possibility, it was made clear that random event space $F \subseteq 2^{\Omega}$ and fuzzy event space $\Sigma \subseteq 2^{\Psi}$ are indeed different in their structures though events in $F \subseteq 2^{\Omega}$ and $\Sigma \subseteq 2^{\Psi}$ are both not mutually exclusive.

*Remark*: In the area of logics and cognition psychology, the topic on the disjunction of natural concept remains open [2,3,17,18]. By psychology experiments, Hampton observed a phenomenon of underextension for disjunction, where underextension means "lower than the maximum" instead of "not lower than the maximum" as required by the maximum rule of fuzzy set disjunction [2,3,18]. Further discussion on these works is out of the scope of this paper.

### *5.4 Axiomatic definition of possibility*

For existing axiomatic definition of possibility, the max operation does not require the condition of being projection-nested. Recall that for the normalized possibility, $\pi(x_i) = 1$ holds for at least one of the elements of *X*. Then the following Proposition 5.1 and Proposition 5.2 hold.

**Proposition 5.1.** In the general case of being not projection-nested, "max" is the only but un-strict disjunctive operator that is applicable across the fuzzy event space.

*Proof*: From (30), we know "max" is the lower bound of the disjunctive operation for calculating $\pi(x_i \cup x_j)$. Suppose there exists operator "$O$", which is other than "max", so that $\pi(x_i \cup x_j) = O(\pi(x_i), \pi(x_j))$. Since $\exists x_i \subseteq \Psi$ so that $\pi(x_i) = 1$, then $\pi(x_i \cup x_j) = O(\pi(x_i), \pi(x_j)) > \max(\pi(x_i), \pi(x_j)) = 1$, which violates the demand that the maximum value of possibility over the sample space must be one. Hence Proposition 5.1 holds.

**Proposition 5.2.** In the general case of being not projection-nested, (34) below holds and "max" is an exact operator for fuzzy feature extraction, i.e., extracting the value from the fuzzy sample space that leads to the largest possibility of one.

$$\pi(\Psi) = \max_{x_i} \pi(x_i) = 1. \qquad (34)$$

*Proof*: Suppose $\Psi = \cup_{i=1}^{N} x_i$. Since $\exists x_i \subseteq \Psi$ so that $\pi(x_i) = 1$, then $\pi(\Psi) = \pi(\cup_{i=1}^{N} x_i) = 1 = \max_{i=1}^{N} \pi(x_i)$.

In Definition 5.1, the axiomatic definition of probability was developed under the condition of being mutually exclusive, which can be fully satisfied among the random sample space. Nevertheless, the condition of being projection-nested in general cannot be satisfied among the fuzzy sample space. Therefore, it would be too strong to require the condition of being projection-nested to be satisfied for the developing of the axiomatic definition of possibility, which would make it very hard to be used for practical applications. On the other hand, Propositions 5.1 and 5.2 tell us that in the general case of being not projection-nested "max" is the only but un-strict disjunctive operator that is applicable across the fuzzy event space, and "max" is an exact operator for extracting the value from the fuzzy sample space that leads to the largest possibility of one. Therefore, it should be a good trade-off to relax the condition of being projection-nested, which results in the well-known definition of possibility below as in [40,46].

**Definition 5.2.** *Possibility* (axiomatic definition) can be built up upon a possibility space $(\Psi, \Sigma, \Pi)$, which is a mathematical construct that models a real-world process of simultaneously classifying an object into multiple fuzzy concepts. A possibility space consists of three parts:

1) the fuzzy sample space $\Psi$, as defined in Definitions 3.6 and 3.7.

2) the σ-algebra $\Sigma \subseteq 2^{\Psi}$, which is the event space consisting of a set of events $\{x_i\}$.

3) the possibility measure $\Pi: \Sigma \longrightarrow [0,1]$, which is a function on $\Sigma$ such that it satisfies the three axioms below:

Axiom 1. (Nonnegativity Axiom) $\pi(\phi) = 0$ for empty set $\phi$.

Axiom 2. (Normality Axiom) the measure of entire fuzzy sample space is equal to one: $\pi(\Psi) = 1$.

Axiom 3. (Maxitivity Axiom) for every countable sequence of events $\{x_i\}$, we have

$$\pi(\cup_{i=1}^{\infty} x_i) = \max_{x_i} \pi(x_i) \qquad (35)$$

By replacing Axiom 3 with Axiom 3' below, we suggest in this work the exact axiomatic definition of possibility satisfying Axiom 1, Axiom 2, and Axiom 3'.

Axiom 3'. (Exact maxitivity Axiom) for every countable sequence of projection-nested events $\{x_i\}$, we have (35).

*Remark*: For both the axiomatic definition and the exact axiomatic definition, "max" is an exact operator as indicated by (34) for extracting the value from the fuzzy sample space that leads to the largest possibility of one. Such a delightful property is very important for practical applications of possibility, which makes sure that max inference discussed below stands as an exact mechanism.

## 6. The induced sigma-max system

We focus on typical forms of sigma-max inference, which consist of composition of uncertain relations and uncertainty update. These two typical forms of uncertainty inference are widely used in the areas of state estimation, target recognition and fuzzy logic system, etc. We mention that Definitions 5.1 and 5.2 of probability and possibility are both established upon the denumerable space, though they have already been discussed into the continuous space [27,46]. To make it concise, our mathematical formulation on sigma-max inference will be limited to discrete variable, but discussions on the application of sigma-max inference will cover the case of continuous variable.

*6.1 The sigma system*

By Axiom 2 and 3 of Definition 5.1, we can derive

$$p(\Omega) = \Sigma_{i=1}^{N} p(x_i) = 1, \qquad (36)$$

which reveals the nature of random variable extraction.

*6.1.1 Conditional probability*

Suppose $p(x_i y_j)$ is the joint probability distribution of random variables *X* and *Y*, then conditional probabilities $p(y_j|x_i)$ and $p(x_i|y_j)$ are defined as

$$p(x_i y_j) = p(y_j|x_i)p(x_i) = p(x_i|y_j)p(y_j), \qquad (37)$$

where

$$p(x_i) = \Sigma_{y_j} p(x_i y_j),\ p(y_j) = \Sigma_{x_i} p(x_i y_j), \qquad (38)$$

which are the equations that are always used for random variable extraction in deriving sigma inference.

*6.1.2 Composition of random relations*

Random relation can be expressed by conditional probability. Suppose $p(y_j|x_i)$ and $p(z_k|y_j)$ represent random relations from *X* to *Y* and from *Y* to *Z*, respectively, then random relation $p(z_k|x_i)$ from *X* to *Z* can be given by

$$p(z_k|x_i) = \sum_{y_l} p(z_k, y_l|x_i) = \sum_{y_l} p(z_k|y_l)p(y_l|x_i), \quad (39)$$

which is derived by extraction of intermediate random variable *Y* from $p(z_k|x_i)$, and $z_k$ and $x_i$ are assumed to be stochastically independent given $y_l$. Eq. (39) indicates that a "sigma-product" combination of two random relations derives a combined random relation.

*6.1.3 Probability update*

Probability update takes the well-known Bayesian form of (40) below

$$p(x_i|y_j) = \frac{p(x_i)p(y_j|x_i)}{p(y_j)} = \frac{p(x_i)p(y_j|x_i)}{\sum_{x_k} p(x_k)p(y_j|x_k)}, \qquad (40)$$

where $p(x_i|y_j)$ is posteriori probability, $p(x_i)$ is priori probability, and $p(y_j|x_i)$ is probability likelihood of $x_i$.

*6.2 The max system*

In parallel to (36), Eq. (34) reveals the nature of fuzzy variable extraction. As claimed previously in Proposition 5.2, "max" is an exact operator for extracting the value from the fuzzy sample space that leads to the largest possibility of one, no matter the condition of projection-nested holds or not. Therefore, starting from (34), all the equations of max inference discussed below stand as exact solutions with no approximation.

*6.2.1 Conditional possibility*

Suppose $\pi(x_i y_j)$ is the joint possibility distribution of fuzzy variables $X$ and $Y$, then conditional possibilities $\pi(y_j|x_i)$ and $\pi(x_i|y_j)$ can be defined using the product rule below [40]

$$\pi(x_i y_j) = \pi(y_j|x_i)\pi(x_i) = \pi(x_i|y_j)\pi(y_j)\ , \qquad (41)$$

where

$$\pi(x_i) = \max_{y_j}\pi(x_i y_j),\ \pi(y_j) = \max_{x_i}\pi(x_i y_j)\ , \qquad (42)$$

which are the equations that are always used for fuzzy variable extraction in deriving max inference. Note that on continuous numerical universe, conditional possibility based on minimum operator (instead of product operator) will induce undesirable discontinuities and the maxitivity axiom will not be preserved [53].

*6.2.2 Composition of fuzzy relations*

Fuzzy relation can be represented by conditional possibility, since membership function of fuzzy sets is equal to conditional possibility with likelihood expansion. Suppose $\pi(y_j|x_i)$ and $\pi(z_k|y_j)$ represent fuzzy relations from $X$ to $Y$ and from $Y$ to $Z$, respectively, then fuzzy relation $\pi(z_k|x_i)$ from $X$ to $Z$ can be given by [31]

$$\pi(z_k|x_i) = \max_{y_l}\pi(z_k, y_l|x_i) = \max_{y_l}\pi(z_k|y_l)\pi(y_l|x_i)\ , \quad (43)$$

which is derived by extraction of intermediate fuzzy variable $Y$ from fuzzy relation $\pi(z_k|x_i)$, and $z_k$ and $x_i$ are assumed to be independent given $y_l$. As we can see from (43), composition of two fuzzy relations equals to the "max-product" operation of two conditional possibilities.

*6.2.3 Possibility update*

From (41) and (42) we can derive a possibility update equation [28,31] in a form parallel to Bayesian inference as

$$\pi(x_i|y_j) = \frac{\pi(x_i)\pi(y_j|x_i)}{\pi(y_j)} = \frac{\pi(x_i)\pi(y_j|x_i)}{\max_{x_k}\pi(x_k)\pi(y_j|x_k)}, \qquad (44)$$

where $\pi(x_i|y_j)$ is posteriori possibility, $\pi(x_i)$ is priori possibility, and $\pi(y_j|x_i)$ is possibility likelihood of $x_i$.

*6.3 A unified statistical interpretation of sigma-max system*

We managed to derive the "sigma" and "max" disjunctive operations from the intuitive definitions of probability and possibility, respectively, by which axiomatic definitions are established. Then how should we make the choice of them for practical applications. We will leave it until after we give a unified statistical interpretation for the sigma-max system. The statistical interpretation of sigma system can straightforwardly attribute to the frequency definition of probability as formulated by (12). Such a statistical characteristics can also be found for max system by the intuitive definition of possibility of (15), where, e.g., size $|f_{x_i}|$ can be recognized as a statistic of the intension $f_{x_i}$. Besides, as discussed in Section 2.2, $f_{x_i}$ can be acquired through statistical learning. Let us make this point clearer using an example of statistical experiment below.

**Example 6.1.** Consider a statistical experiment of investigating public opinion $X$ on a recent social event. Suppose you as the investigator have collected 100 comments released on free media, and you designed a Table 1 to make statistical analysis. You will fill in the table with judgment whether the public opinion is POSITIVE $x_1$, NEUTRAL $x_2$, or NEGATIVE $x_3$. In filling Table 1 you find many of the collected comments are quite "fuzzy", which means these comments may simultaneously contain, e.g., positive and negative words, you decide to design another Table 2 to manipulate it.

Table 1: You need to vote only one of the outcomes (POSITIVE, NEUTRAL, or NEGATIVE) for one of the 100 comments, which means this is an either-or choice. By definition, the uncertainty involved should be recognized as randomness and could be measured by probability. And we got the vote results as in Table 1.

Table 2: You are allowed to vote 1~3 choices from the outcomes (POSITIVE, NEUTRAL, or NEGATIVE) simultaneously for one of the 100 comments, which means this is a both-and choice. By definition, the uncertainty involved should be recognized as fuzziness and could be measured by possibility. And we got the vote results as in Table 2.

As we can see in Table 1, among 100 comments, you choose 60 votes POSITIVE, 25 votes NEUTRAL and 15 votes NEGATIVE. Therefore, the public opinion is POSITIVE with probability of 0.6, NEUTRAL with probability of 0.25, and NEGATIVE with probability of 0.15. In Table 2, POSITIVE has 90 votes whereas NEUTRAL has 45 votes, and NEGATIVE has 18 votes. Therefore, the public opinion is POSITIVE with possibility of 1, NEUTRAL with possibility of 0.5, and NEGATIVE with possibility of 0.2. Here 45 votes of NEUTRAL, e.g., can be regarded as an evaluation of size of the intension of public opinion $X$ being NEUTRAL $x_2$, i.e., $|f_X \cap f_{x_2}|$. And $|f_X|$ has to be 90 votes instead of 100 votes, because definition (15), as normalized possibility, requires that $|f_X| = |f_X \cap f_{x_i}|_{max}$. In fact, the mysterious $|f_X|$ can never be set and can only be dug out through the evaluation process of $|f_X \cap f_{x_i}|$. And these possibilities can be transformed into probabilities by using (45) or the method shown in Table 2. Probability can be transformed into possibility using (46), as well [40].

$$p(x_i) = \frac{\pi(x_i)}{\sum_j \pi(x_j)} \qquad (45)$$

$$\pi(x_i) = \frac{p(x_i)}{\max_j p(x_j)} \qquad (46)$$

TABLE 1. VOTE RESULTS WITH EITHER-OR CHOICE

| | **POSITIVE** $(x_1)$ | **NEUTRAL** $(x_2)$ | **NEGATIVE** $(x_3)$ |
|---|---|---|---|
| **votes** | 60 | 25 | 15 |
| **probability** | 60/100=0.60 | 25/100=0.25 | 15/100=0.15 |

TABLE 2. VOTE RESULTS WITH BOTH-AND CHOICE

| | **POSITIVE** $(x_1)$ | **NEUTRAL** $(x_2)$ | **NEGATIVE** $(x_3)$ |
|---|---|---|---|
| **votes** | 90 | 45 | 18 |
| **possibility** | 90/90=1.00 | 45/90=0.50 | 18/90=0.20 |
| **transformed probability** | 90/153=0.59 | 45/153=0.29 | 18/153=0.12 |

*Remark*: In fact, the public opinion $X$ on the investigated event is neither random nor fuzzy, but is unknown to us and would have a certain spread or distribution across the public. Nevertheless, the uncertainty involved in the collected 100 comments can be handled as either randomness or fuzziness, and by two different statistical ways – vote with either-or choice, or vote with both-and choice, which hence produces different forms of distribution over the outcomes of POSITIVE, NEUTRAL, and NEGATIVE. Another typical example related may be a Fallen Cup. A cup falling to the floor will randomly break or not break, but a fallen cup on the floor without observation is neither random nor fuzzy but unknown to us. Nevertheless, we can make sigma or max inference to the status of the fallen cup according to available knowledge and/or data. In a more general sense, Example 6.1 tell us that for the task of uncertain inference, interpretation of the unknown uncertainty relies on the available prior information and the perspective of modeling. The uncertainty involved in the unknown quantity, e.g., the public opinion $X$ on the investigated event, can be interpreted as either randomness or fuzziness as shown in Fig. 9 [34]. Under the perspective of occurrence of objective event, the unknown public opinion $X$ should take place as an objective event $x_i$ from the set of exclusive outcomes. In such a case, $X$ should be interpreted as randomness. Under the perspective of subjective cognition of concept (event), the unknown public opinion $X$ can be simultaneously classified into more than one outcome from the set of non-exclusive outcomes. In such a case, $X$ should be interpreted as fuzziness. The fuzzy uncertainty arisen here has common in essence with that of the natural fuzzy concept such as YOUTH, i.e., fuzziness is caused by the overlap of their intensions. Though the random-fuzzy dual interpretation of uncertain variable is advocated above, we in Section 6.4 will give some guidance for the choice of sigma-max inference, which have been verified by some experimental examples.

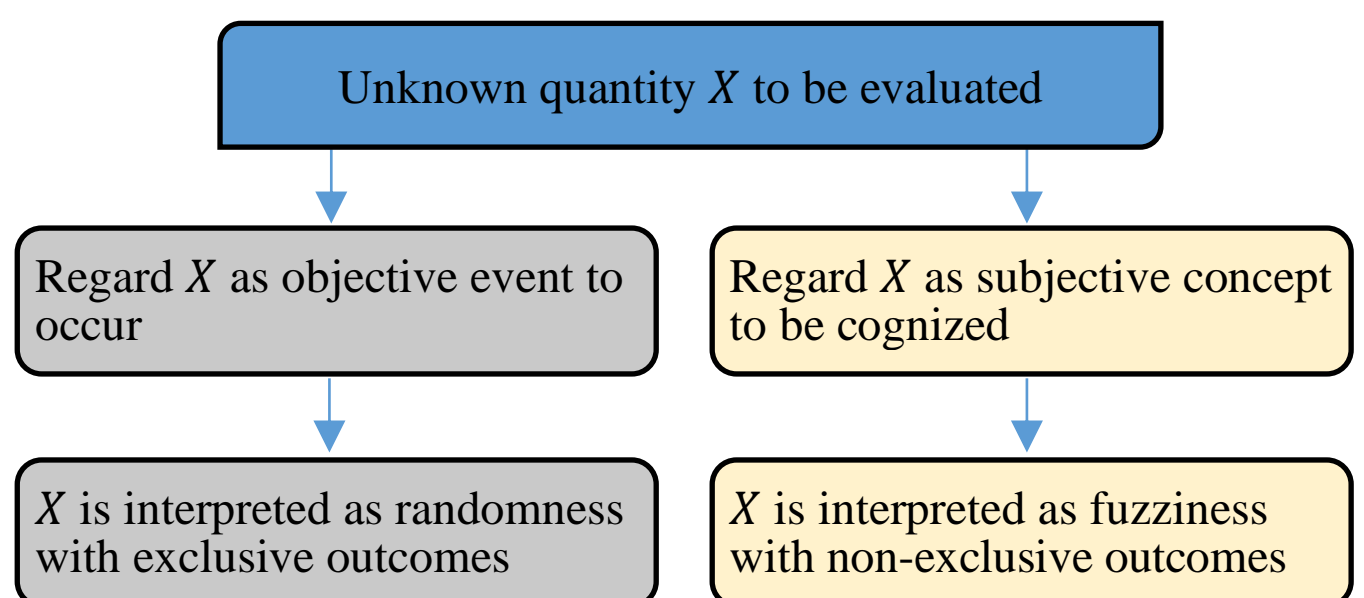

Fig. 9. The random-fuzzy dual interpretation of uncertain variable.

### *6.4 On the choice of sigma-max system*

From the perspective of this work, the operators of sigma and max were born in randomness and fuzziness, respectively. Nevertheless, it would be not appropriate to simply equate sigma operator with randomness, and max operator with fuzziness. The logic behind is that from the intuitive definition $\boldsymbol{A}$ we derived axiomatic definition $\boldsymbol{B}$, mathematically $\boldsymbol{A} \Longrightarrow \boldsymbol{B}$, but this does not mean $\boldsymbol{A} = \boldsymbol{B}$. The sigma (max) operator reflects some representative and probably intrinsic but not all characteristics of the random (fuzzy) uncertainty in terms of disjunctive operation. Once born, the sigma (max) operator would have its own life. For the choice of uncertainty theory for practical applications, recognition of the uncertainty involved is no doubt of key importance. Nevertheless, functions built in with the "sigma" and "max" should be the direct reference factors. We take the perspective that:

*a) Discrete variable involved in statistical inference, to which the procedure of feature extraction usually applied, could in general be better modeled as fuzziness.*

*b) Continuous variable involved in statistical inference, to which the procedure of feature extraction usually does not apply, should in general be regarded as randomness.*

The process of feature extraction causes the overlap of the extracted intensions hence the fuzziness of the discrete variable being inferred. In many applications of statistical inference, we collect data from various kinds of sensors, which in most cases are continuous quantities. Measurement data from sensor are usually mixed with noise, and the involved measurement uncertainty itself should be regarded as randomness and is usually evaluated on probability distributions [51].

For a detailed discussion on the choice of sigma and max operators by considering their build-in functions, the readers may refer to [34], which is consistent with the viewpoint that sigma and max operators are suitable for coping with continuous and discrete variables, respectively. The key difference of the two operators lies in the fact that during the process of uncertainty inference, sigma operator uses all possible values of the intermediate variable of $Y$, whereas max operator selects only one value that would make the combined possibility $\pi(z_k|x_i)$ having the maximum of one. This property of possibility originates from (34) which claims that at least one of the elements of $\Psi$ should be fully possible.

The view above on the choice of sigma-max system reminds us of two opposite positions on the relationship between mathematics and the physical world. A school of thought, reflecting the ideas of Plato, is that mathematics has its own existence; whereas the opposing viewpoint is that mathematical forms are objects of our human imagination and we make them up as we go along, tailoring them to describe reality [1].

### *6.5 Towards an integrated sigma-max system*

The sigma-max system discussed above are two parallel uncertainty systems appropriate for handling randomness and fuzziness, respectively. Nevertheless, in many cases we will come across random uncertainty and fuzzy uncertainty simultaneously. For example, in some applications of target recognition, we are aware that the uncertainty related to target type and feature could be more appropriately modeled as fuzzy variables whereas the uncertainty related to observation should be regarded as random variable. For such cases, a joint description of heterogeneous uncertainty should be a desired solution, and the separate uncertainty systems of probability and possibility need to be developed into an integrated sigma-max system [32]. To help present the example in Section 7, we below summarize some key points of the integrated system. We suggest to use sigma-max system (inference) to refer to both the integrated sigma-max system and the parallel sigma/max systems.

By defining hybrid distribution $p\pi(x_i y_j)$, or $\pi p(y_j x_i)$ of a random variable *X* and a fuzzy variable *Y*, two groups of equations for hybrid uncertainty inference are derived, which include composition of heterogeneous relations (47) and (48), and uncertainty update with heterogeneous information (49) and (50).

#### *6.5.1 Composition of heterogeneous relations*

$$p(x_i|y_j) = \frac{1}{\beta} p^+(x_i|y_j) = \frac{1}{\beta} \max_{z_k} p(x_i|z_k)\pi(z_k|y_j) \quad (47)$$

$$\pi(y_j|x_i) = \frac{1}{\beta} \pi^+(y_j|x_i) = \frac{1}{\beta} \Sigma_{s_k} \pi(y_j|s_k)p(s_k|x_i) \quad (48)$$

where $p(x_i|y_j)$ denotes the probability of random variable *X* being $x_i$ given fuzzy variable *Y* being $y_j$, and $\pi(y_j|x_i)$ is the possibility of $y_j$ conditioned on $x_i$, similar denotations hold for other expressions. $p^+(x_i|y_j)$ and $\pi^+(y_j|x_i)$ are induced distributions and $\beta$ is a normalization factor such that $p(x_i|y_j)$ and $\pi(y_j|x_i)$ are probability distribution and possibility distribution, respectively.

The induced $p^+(x_i|y_j)$ is a direct, hence accurate, fusion of the prior knowledge of $p(x_i|z_k)$ and $\pi(z_k|y_j)$. By a normalization factor of $\beta$, we can have $p(x_i|y_j)$, the probability version of $p^+(x_i|y_j)$. Be aware both $p^+(x_i|y_j)$ and $p(x_i|y_j)$ are many-to-many mappings, which can be either distributions over different $x_i$s conditioned on a certain value of $y_j$, or likelihood functions of different $y_j$s given a certain value of $x_i$. In the former case, it would be necessary to transform $p^+(x_i|y_j)$ into $p(x_i|y_j)$. In the latter case, such a conversion would potentially introduce normalization bias and is unnecessary since $\beta$ in general has different values for different $y_j$. Similarly, the application of (48) should make clear whether we need a likelihood expansion of $\pi^+(y_j|x_i)$ over $x_i$s or a distribution of $\pi(y_j|x_i)$ over $y_j$s.

#### *6.5.2 Uncertainty update with heterogeneous information*

Eq. (49) is possibility update equation with prior possibility $\pi(y_j)$ and probability likelihood $p(x_i|y_j)$.

$$\pi(y_j|x_i) = \frac{p(x_i|y_j)\pi(y_j)}{\max_{y_l} p(x_i|y_l)\pi(y_l)} \quad (49)$$

Eq. (50) is probability update equation with prior probability $p(x_i)$ and possibility likelihood $\pi(y_j|x_i)$.

$$p(x_i|y_j) = \frac{\pi(y_j|x_i)p(x_i)}{\sum_{x_k}\pi(y_j|x_k)p(x_k)} \quad (50)$$

## 7. An example of stock price prediction using real dataset

### *7.1 The task*

This section provides a demonstration example of stock price prediction using real dataset, which confirms that sigma-max inference indeed exhibits distinctive performance than sigma inference for the investigated application. Prediction of stock price belongs to a broader topic of time series analysis, which has attracted much attention in the past three decades [6,20,26]. Typical methods may include autoregressive integrated moving average (ARIMA) methods, state space models, and ANN models. Our method is based on state space model, and we are specifically interested in prediction of a kind of non-stationary time series – stock price prediction [20,50]. We use data of stock price of *Google Inc.* and *Comcast Corp* for the past 2520 working days (approximately 10 years) when the market was open [50]. For our experiments, these data are normalized as shown in Fig. 10. The goal is to predict the closing price of stock for the $(k+1)^{th}$ day ($z_{k+1}$) given the close-price series till $k^{th}$ day ($z_{1:k}$).

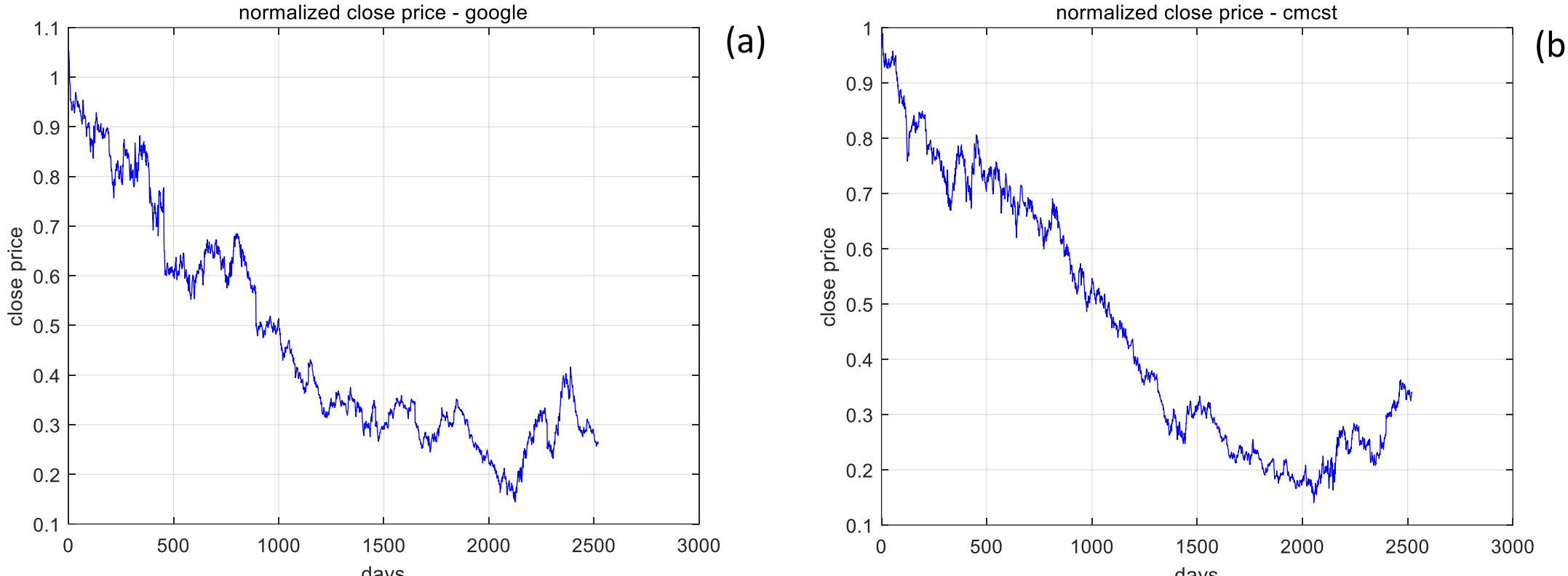

Fig. 10. Normalized closing price for the past 2520 working days. (a) *google*. (b) *cmcst*.

### *7.2 The state space model*

We use the following Markovian jump linear system to model non-stationary time series - the stock price data. The linear model can be extended into a non-linear system if necessary.

$$x_k = F(r_k)x_{k-1} + G(r_k)w_k \quad (51)$$

$$z_k = H(r_k)x_k + v_k \quad (52)$$

where system mode index $r_k$ is described by a finite Markovian chain, which takes values from model set $\mathcal{M} = \{1,2 \ldots, M\}$. $F(r_k)$, $G(r_k)$ and $H(r_k)$ are known matrices for each mode index $r_k$. Transition function $F(r_k)$ models the evolution of continuous state vector $x_k$ at time $k$ as a first-order Markovian process, with modeling error represented by random process noise $w_k \sim \mathcal{N}(0, Q_k)$ weighted through noise gain $G$. Observation function $H(r_k)$ models the relationship between the state $x_k$ and the measurement (closing price) $z_k$ with measurement error denoted by $v_k \sim \mathcal{N}(0, R_k)$.

Above, process noise $w_k$ and measurement noise $v_k$ are modeled as random uncertainty. Traditionally, the transition of system mode is modeled by the transition probability matrix

$$\mathcal{P} = [p_{ij}]_{M\times M} = [p(r_k = j|r_{k-1} = i)]_{M\times M} \quad (53)$$

where $p_{ij}$ denotes the transition probability from mode *i* to mode *j* and it satisfies $\sum_{j=1}^{M} p_{ij} = 1$ for any $i \in \mathcal{M}$. Markovian jump system formulated by (51) ~ (53) is a stochastic linear system.

According to the perspective presented in this paper, the system mode $r_k$, which is an unknown and extracted higher-level discrete process, could be better interpreted as fuzziness and modeled by the transition possibility matrix

$$\Pi = [\pi_{ij}]_{M\times M} = [\pi(r_k = j|r_{k-1} = i)]_{M\times M} \qquad (54)$$

where $\pi_{ij}$ denotes the transition possibility from mode $i$ to mode $j$ and it satisfies $\max_{j\in\mathcal{M}} \pi_{ij} = 1$ for any $i \in \mathcal{M}$. We name the hybrid uncertainty system formulated by (51), (52) and (54) as a fuzzy Markovian jump system, and system of (51) ~ (53) as a random Markovian jump system.

*7.3 The predictor for real dataset*

The celebrated interacting multiple model (IMM) filter was developed for the stochastic linear system (51) ~ (53), to compute the posterior distribution $p(x_k|z_{1:k})$ and/or its first and second moments $\hat{x}_{k|k}$ and $\hat{\Sigma}_{k|k}$ [7,15]. Given the hybrid uncertainty linear system (51), (52) and (54), a sigma-max version of the IMM (SIMM) filter was developed by using the integrated sigma-max inference [34], which has a structure that is parallel to the classic IMM filter. During a circle of the algorithm, the SIMM runs in parallel $M$ model-matched filters and only one filter with the maximum model possibility is fully in charge of the filtering owing to its built-in decision of the system model. Whereas for the IMM, each model-matched filter will share a certain probability of responsibility of the filtering work. We use the classic IMM and the new developed SIMM as predictors for the stock price, and compare their prediction accuracy $\tilde{z}_{k+1|k}$ of the closing price, which is given by

$$\tilde{z}_{k+1|k} = z_{k+1} - \hat{z}_{k+1|k} = z_{k+1} - H\hat{x}_{k+1|k}, \qquad (55)$$

where $z_{k+1}$ is closing price at $(k+1)^{th}$ day, and $\hat{z}_{k+1|k}$ is its predicted price by the predictor using close-price series $z_{1:k}$ till $k^{th}$ day.

*7.4 Experiment setup*

To best cover the dynamics of the stock price data, three dynamic models are selected for both the IMM and the SIMM, which are the static (ST) model, the constant velocity (CV) model and the constant acceleration (CA) model [34]. Covariances of the process/measure noise are evaluated by analyzing the statistics of the dataset. For all three datasets and all three models, we set process noise covariance $Q_k = 0.002^2$, and measurement noise covariance $R_k = 0.002^2$. The three models have equal initial probabilities/possibilities with transition matrices fine-tuned and given by

$$\mathcal{P} = [p_{ij}]_{M\times M} = \begin{pmatrix} 0.950 & 0.025 & 0.025 \\ 0.025 & 0.950 & 0.025 \\ 0.025 & 0.025 & 0.950 \end{pmatrix},$$

$$\Pi = [\pi_{ij}]_{M\times M} = \begin{pmatrix} 1 & 0.2 & 0.2 \\ 0.2 & 1 & 0.2 \\ 0.2 & 0.2 & 1 \end{pmatrix}.$$

*7.5 Experimental results*

Presented in Table 3 are prediction accuracies in terms of root mean squared error (RMSE) obtained by each method on these stock price datasets. The absolute value of prediction errors over time are shown Fig. 11. As can be seen, our method SIMM exhibits significant improvements over the classic IMM. Model probabilities of IMM and SIMM are shown in Figs. 12 and13, where model possibility of SIMM is transformed into probability using (45). As can be seen, particularly from Fig. 12, model transition of SIMM is much faster than IMM, which probably explains why SIMM surpasses IMM for the investigated problem.

## 8. Conclusion

From the intuitive definition of possibility, we derived the disjunctive operation of "max" when the condition of being projection-nested is satisfied. We conclude that in the general case of being not projection-nested, "max" is the only but un-strict disjunctive operator that is applicable across the fuzzy event space, and is an exact operator for extracting the value from the fuzzy sample space that leads to the largest possibility of one, which assures the max inference is an exact mechanism. In contrast, sigma operator is appropriate for probability in disjunctive operation of mutually-exclusive events. Our work provides a physical foundation for the axiomatic definition of possibility for the measure of fuzziness by the introduction of intuitive mathematical definition of possibility. Our demonstration example as well as application insights of sigma-max inference reveal that discrete variable extracted in statistical inference generally could be better modeled as fuzziness and handled by max operator. Hopefully, our work would help facilitate wider adoption of possibility theory in practice, and promote cross prosperity of the two uncertainty theories of probability and possibility. Possible future studies could be conducted along both theoretical and applicational fronts, such as: a) the demonstration of the intuitive definition of possibility through the utilization of a deep ANN classifier; b) the development of possibility-oriented statistic theory; c) more applications and insights of the sigma-max inference in non-stationary stochastic process, neural-symbolic inference, generative model, and beyond.

TABLE 3. EXPERIMENT RESULTS FOR IMM AND SIMM IN TERMS OF RMSE.

| | *google* | *cmcst* |
|---|---|---|
| IMM | 0.0081 | 0.0079 |
| SIMM | **0.0074** | **0.0064** |
| Δ (%) | -8.64% | -18.99% |

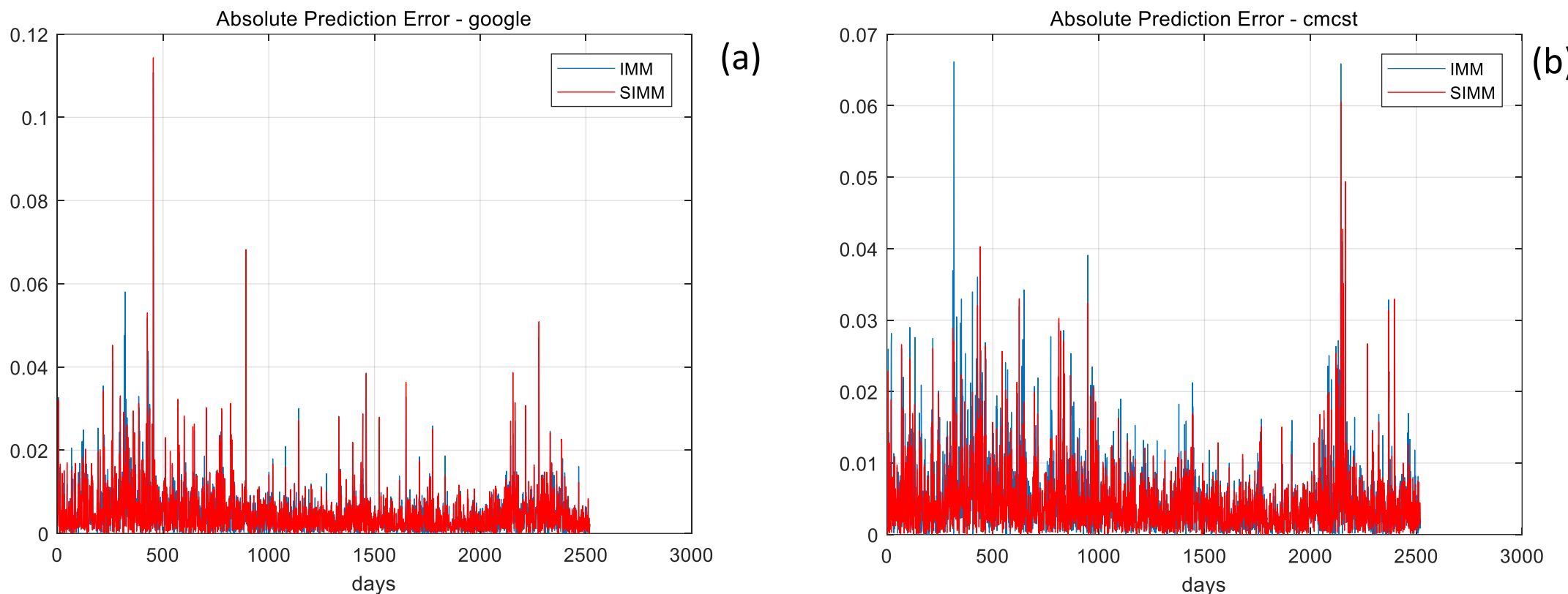

Fig. 11. Absolute prediction error of IMM and SIMM. (a) *google*. (b) *cmcst*.

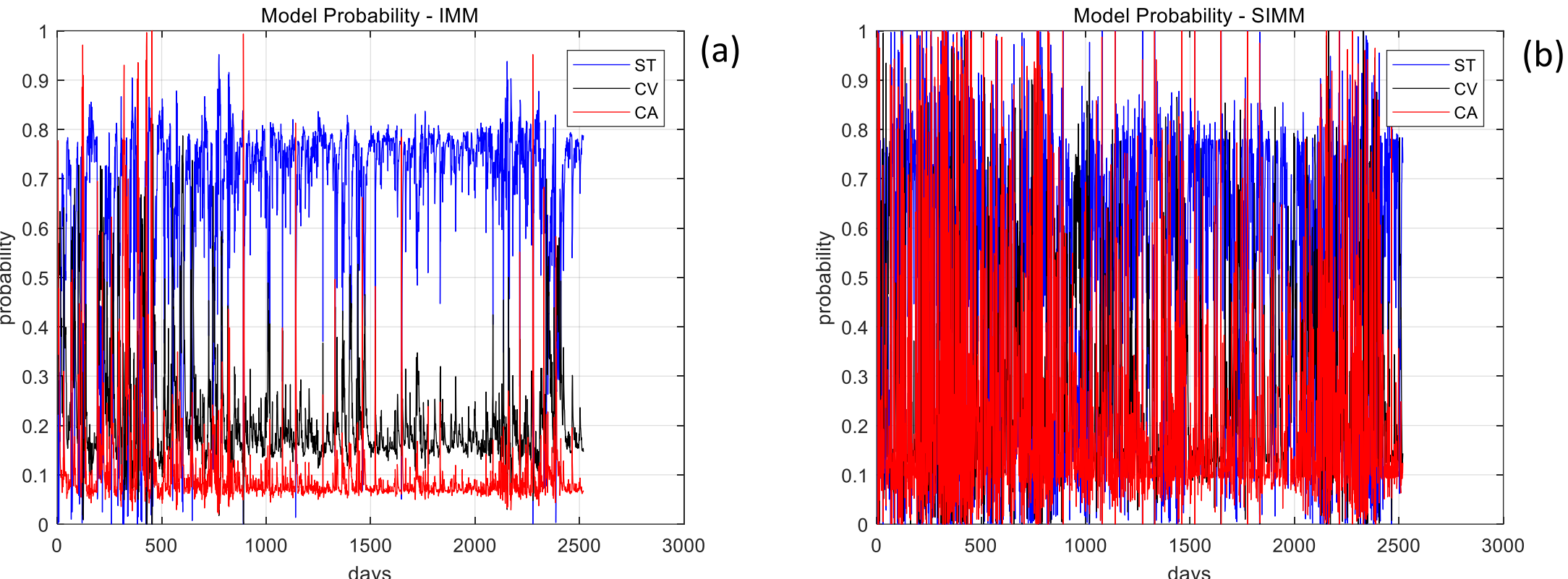

Fig. 12. Model probabilities for *google* data. (a) IMM. (b) SIMM.

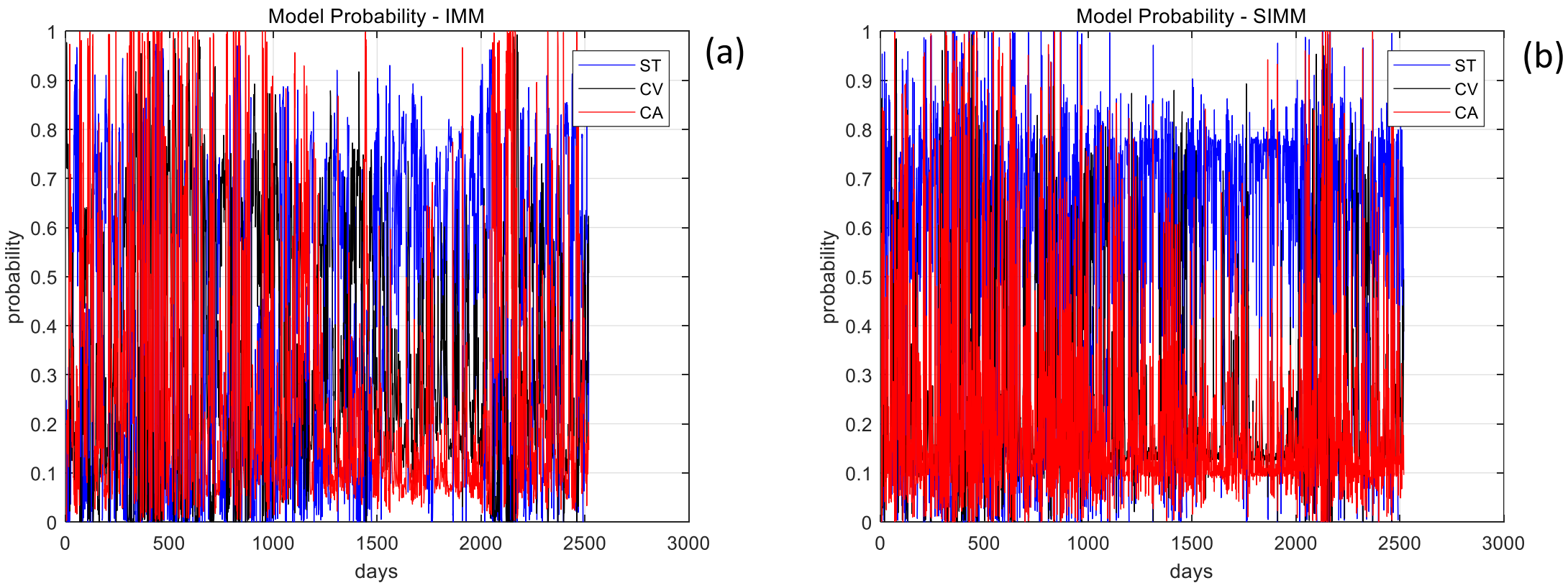

Fig. 13. Model probabilities for *cmcst* data. (a) IMM. (b) SIMM.

## Data availability

Data will be made available on request.